\documentclass[10pt]{article} % For LaTeX2e
\usepackage[accepted]{tmlr}
\usepackage{amsmath,amsfonts,bm}

\def\eqref#1{equation~\ref{#1}}
\def\1{\bm{1}}

\DeclareMathAlphabet{\mathsfit}{\encodingdefault}{\sfdefault}{m}{sl}
\SetMathAlphabet{\mathsfit}{bold}{\encodingdefault}{\sfdefault}{bx}{n}

\usepackage[utf8]{inputenc} % allow utf-8 input
\usepackage[T1]{fontenc}    % use 8-bit T1 fonts
\usepackage{hyperref}       % hyperlinks
\usepackage{url}            % simple URL typesetting
\usepackage{booktabs}       % professional-quality tables
\usepackage{amsfonts}       % blackboard math symbols
\usepackage{nicefrac}       % compact symbols for 1/2, etc.
\usepackage{microtype}      % microtypography
\usepackage{xcolor}         % colors
\usepackage{amsmath}
\usepackage{amssymb}
\usepackage{mathtools}
\usepackage{amsthm}
\usepackage{float} 
\usepackage{bbm}
\usepackage{caption}
\usepackage{subcaption}
\usepackage{tabularx}

\title{What Should Embeddings Embed? Autoregressive Models Represent Latent Generating Distributions}

\author{\name Liyi Zhang \email zhang.liyi@princeton.edu \\
      \addr Department of Computer Science\\
      Princeton University
      \AND
      \name Michael Y. Li \email michaelyli@stanford.edu \\
      \addr Department of Computer Science\\
      Stanford University
      \AND
      \name R. Thomas McCoy \email tom.mccoy@yale.edu \\
      \addr Department of Linguistics and Wu Tsai Institute\\
      Yale University
      \AND
      \name Theodore R. Sumers \email ted@anthropic.com \\
      \addr Anthropic
      \AND
      \name Jian-Qiao Zhu \email jz5204@princeton.edu \\
      \addr Department of Computer Science\\
      Princeton University
      \AND
      \name Thomas L. Griffiths \email tomg@princeton.edu \\
      \addr Departments of Psychology and Computer Science\\
      Princeton University
      }

\def\month{07}  % Insert correct month for camera-ready version
\def\year{2025} % Insert correct year for camera-ready version
\def\openreview{\url{https://openreview.net/forum?id=YyMACp98Kz}} % Insert correct link to OpenReview for camera-ready version

\begin{document}

\maketitle

% TMLR requires electronic submissions, processed by
% \url{https://openreview.net/}. See TMLR's website for more instructions.

% If your paper is ultimately accepted, use option {\tt accepted} with the {\tt tmlr} 
% package to adjust the format to the camera ready requirements, as follows:
% \begin{center}  
%   {\tt {\textbackslash}usepackage[accepted]\{tmlr\}}. 
% \end{center}
% You also need to specify the month and year
% by defining variables {\tt month} and {\tt year}, which respectively
% should be a 2-digit and 4-digit number. To de-anonymize and remove mentions
% to TMLR (for example for posting to preprint servers), use the preprint option,
% as in {\tt {\textbackslash}usepackage[preprint]\{tmlr\}}. 

\begin{abstract}
  Autoregressive language models have demonstrated a remarkable ability to extract latent structure from text. The embeddings from large language models have been shown to capture aspects of the syntax and semantics of language. But what {\em should} embeddings represent? We show that the embeddings from autoregressive models correspond to predictive sufficient statistics. By identifying settings where the predictive sufficient statistics are interpretable distributions over latent variables, including exchangeable models and latent state models, we show that embeddings of autoregressive models encode these explainable quantities of interest. We conduct empirical probing studies to extract information from transformers about latent generating distributions. Furthermore, we show that these embeddings generalize to out-of-distribution cases, do not exhibit token memorization, and that the information we identify is more easily recovered than other related measures. Next, we extend our analysis of exchangeable models to more realistic scenarios where the predictive sufficient statistic is difficult to identify by focusing on an interpretable subcomponent of language, topics. We show that large language models encode topic mixtures inferred by latent Dirichlet allocation (LDA) in both synthetic datasets and natural corpora. 

\end{abstract}

\section{Introduction}

% I. Problem: LLMs are effective. But explainability is lacking. Embeddings are often used for downstream tasks. We need to know what embeddings represent.

Autoregressive language models (LMs) are trained to predict the next token in a sequence \cite[e.g.,][]{bengio2000neural}. Many large language models (LLMs) use the autoregressive objective for pretraining \cite[e.g.,][]{radford2019language}, and their document-level embeddings have been shown to capture elements of latent structure that appear in text, such as agent properties \citep{andreas-2022-language} and syntax \citep{hewitt2019structural}. While it is intuitive that representing this information assists in next token prediction, we lack a formal understanding of which aspects of text should be represented and why. Given that these embeddings are often used for downstream tasks such as sequence classification and sentiment analysis, understanding what they represent and why is key to interpreting and building on these models. %, and to supporting transparency and safety.
%

% II. Challenge: 
Analyzing the representations formed by LLMs is challenging partly due to polysemanticity, where a neuron may activate for several distinct concepts \citep{Cunningham2023SparseAF}. Previous work has developed methods for probing LLM internal representations for specific concepts \citep{Li2023HowDT, Meng2022TopicDV,zheng2023revisiting,tenney2018what}.
These efforts can be guided more effectively by developing a general theory of what aspects of the data embeddings should represent.

% III. Solution: 
% [mention the paper that shows meta-trained RNNs to be Bayes-optimal. We are motivated by Bayes-optimality in LMs. Introduce our method. Summarize contributions: narrowing down on what LM embeddings should represent; showing that LMs represent tasks like Bayes-optimal agents.]

In this work, we investigate several cases where the representations of autoregressive LMs can be formally connected with those of a Bayes-optimal agent. Because of their expressive architecture and their objective, LM embeddings should encode latent structure such that the next word $x_{n+1}$ is independent from previous words $x_{1:n}$ when conditioned on that structure. This corresponds to the notion of a {\em predictive sufficient statistic}. Using this connection, we show that optimal content of embeddings can be identified in two major cases: 1) independent identically distributed data, where the embedding should capture the sufficient statistics of the data; 2) latent state models, where the embedding should encode the posterior distribution over the next latent state given the data. An application of case 1 that we explore is topic models based on latent Dirichlet allocation (LDA), where the LM embedding encodes the topic mixtures of a given document. We use probing to confirm that the relevant information can be decoded from LM embeddings, and that content that is not expected to be captured via predictive sufficient statistics is more challenging to recover.

Next, we extend our analysis of exchangeable models to more realistic scenarios by generating synthetic data where only a fraction of words are generated from LDA. We confirm that language models represent topic distributions when a sufficiently large portion of the text serves this semantic component. 
We finally show that the encodings of pretrained LLMs (\textsc{Gpt-2}, \textsc{Llama 2}, and \textsc{Bert}) contain information analogous to that extracted by LDA from two natural corpora, supporting the hypothesis that LLMs implicitly perform Bayesian inference in the naturalistic setting. 
%Our results suggest that this is because topics form a sub-component of language --- enough semantically-related words can be modeled by a topic model such that topic information is relevant to generating text. 

% IV. Insight: 

Our analysis provides concrete examples where we can analytically identify what embeddings should represent, and empirical confirmation that this information is indeed captured in models trained with an autoregressive objective. 
%These results indicate that certain information will be easier to decode from LMs -- for example, previous work on belief state inference in transformers \citep{shai2024transformersrepresentbeliefstate} focuses on decoding a different conditional distribution from the one identified by our analysis, and we show that decoding performance is worse for this distribution.  
By linking LM representations to Bayesian inference, our approach supports recent arguments that the behavior of LMs can be interpreted through comparison with Bayes-optimal agents \citep{Mikulik2020MetatrainedAI}. Bayesian inference encourages the agent to summarize data by constructing a generative process. Thus, understanding this connection can help LLM researchers hypothesize what features are encoded in LLM internals as done in \textit{mechanistic interpretability} \citep{Cunningham2023SparseAF}; by recovering latent generating distributions from LM, we show that LMs not only learn concepts, but also learn their uncertainty representation, complementing recent studies at the behavioral level \citep{gruver2024largelanguagemodelszeroshot}.

\begin{figure}[t]
    \centering
    \includegraphics[width=0.8\textwidth]{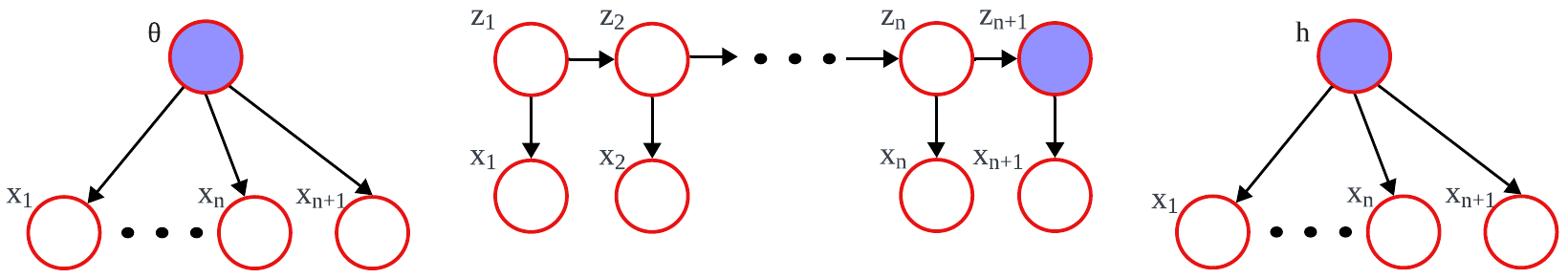}
    \caption{Three data generation processes where prediction of the next token $x_{n+1}$ is independent from previous tokens $x_{1:n}$ given a predictive sufficient statistic. The left corresponds to exchangeable data, the middle to latent state models, and the right to discrete hypotheses. The relevant predictive sufficient statistics are the sufficient statistic for $\theta$, $z_{n+1}, h$
    (or $p(\theta|x_{1:n})$, $p(z_{n+1}|x_{1:n})$, and $p(h|x_{1:n})$ respectively). We show the embeddings learned by autoregressive transformers represent this information.}
    \label{fig:graphical-model}
\end{figure}

\section{Related work}

\paragraph{Embedding analysis in language models} The embeddings produced by language models have been investigated in detail \citep{gupta2015distributional,kohn2015whats,ettinger2016probing,adi2017finegrained,hupkes2018visualisation}; for reviews, see \citet{rogers2020primer} and \citet{belinkov2022probing}. 
% They have been shown to capture aspects of the latent structure of text, including part of speech \citep{shi2016string,belinkov2017neural}, syntactic number \citep{giulianelli2018hood,conneau2018cram}, sentence structure \citep{tenney2018what,hewitt2019structural,liu2019linguistic,lin2019open}, entity attributes \citep{gupta2015distributional,grand2022semantic}, sentiment \citep{radford2017learning}, semantic roles \citep{ettinger2016probing,tenney2018what}, world states \citep{li2021implicit}, and agent properties \citep{andreas-2022-language}. 
They have been shown to capture aspects of the latent structure of text, including part of speech \citep{shi2016string,belinkov2017neural},  sentence structure \citep{tenney2018what,hewitt2019structural,liu2019linguistic,lin2019open}, sentiment \citep{radford2017learning}, semantic roles \citep{ettinger2016probing,tenney2018what}, and agent properties \citep{andreas-2022-language}. 
However, our work is motivated from a fundamentally different perspective. Instead of focusing on \textit{what} is captured in the embeddings of these models, we provide a general theory of language models as Bayes-optimal agents to explain \textit{why} these kinds of structure might be represented due to statistical properties of the training data.
Characterizing the ideal representations of a Bayes-optimal agent
complements recent efforts in \textit{mechanistic interpretability} \citep{nanda2023progress, Cunningham2023SparseAF, cammarata2020thread:} by identifying optimal representations of data; this can inform the search for interpretable features that are used in LLMs' circuit-level computations. 
% TLG: probably don't need this full list, could just include these into the general citations above or say they have focused on different aspects of language

\paragraph{Implicit Bayesian inference} Several previous papers have analyzed LLMs by making a connection to Bayesian inference. Of these, \citet{xie2021explanation}, \citet{mccoy2023embers}, and \citet{wang2023large} analyze the in-context learning behavior of LLMs. However, we study what models should encode based on the autoregressive objective that is typically used to train LLMs. \citet{zheng2023revisiting} focus on topic models embedded in LSTMs, while we extend the connection to more general cases. \citet{wang-etal-2024-variational} provides a novel method to extract from language models concepts that come in the form of a mixed membership model (including the topic model). Our focus is more general and explanatory as we analyze why concepts that come in the form of sufficient stats should be embedded by autoregressive language models.

Belief state inference (BSI) on latent state models \citep{shai2024transformersrepresentbeliefstate} decodes the distributions of the current latent state from the transformer embeddings on observed sequences. Our more general theory shows that this ability comes from the embedding encoding predictive sufficient statistics for the sequence, which more directly relates to the distribution of the next latent state given observations.  Concurrently, \citet{ye2024pretrainingincontextlearningbayesian} also used de Finetti's theorem to support probabilistic reasoning in transformers, but focused on {\em in-context learning} in the context of Bayesian linear regression. We analyze transformer internal representations and identify optimal embeddings in data that violate the exchangeability assumption.

\paragraph{Theory} \citet{Tishby2015DeepLA} gave a theoretical analysis of optimal embeddings in deep networks from an information-theoretic perspective. We take a complementary Bayesian approach, explicitly connecting optimal embeddings to predictive sufficient statistics and presenting extensive experiments confirming that the predicted content of embeddings is tracked by transformer-based LMs. Metalearned RNNs have also been shown to encode information equivalent to a Bayesian posterior distribution \citep{Mikulik2020MetatrainedAI}. 
Furthermore, recent work has also demonstrated that transformers behave like the Bayes-optimal predictor in linear regression settings \citep{panwar2024incontext, garg2022what, akyrek2023what} and can approximate the posterior predictive distributions of probabilistic models such as Gaussian processes and Bayesian neural networks \citep{Muller2022transformers}.
We extend this analysis to general autoregressive language models and consider more general generative processes and what posterior distributions they should capture in these cases.

% TLG: we need more stuff on NNs representing Bayes, maybe check in this paper for references?

\section{Optimal embeddings}

% \michael{Might be worth thinking of a good figure 1...}

Assume we have a sequence $x_{1:n}$ and an autoregressive language model that predicts the next item in the sequence, $p(x_{n+1}|x_{1:n})$. We denote the LM embedding for sequence $x_{1:n}$ as $\phi_n = f(x_{1:n})$. The distribution $p(x_{n+1}|x_{1:n})$ is a function $g(\phi_n)$ of this embedding, with that function implemented by the final layer of the neural network instantiating the LM such that the probability of the next element $x_{n+1}$ only depends on $\phi_n$. This establishes our question: what should $\phi_n$ represent in order to accurately predict $x_{n+1}$?

The key idea behind our approach is that we can identify situations where  $\phi_n$ contains all of the information from $x_{1:n}$ required to predict $x_{n+1}$ by using the notion of a {\em sufficient statistic} \citep{gelmanbda04}.
Given a distribution $p(x)$ with parameters $\theta$, a statistic $s(x)$ is sufficient for $\theta$ if the conditional distribution of $x$ given $s$ does not depend on $\theta$. In other words, if we only know $s$, we can estimate $\theta$ just as well as if we know $x$. 
% For example, if $x$ is a set of independently and identically distributed draws from a Gaussian distribution and $\theta$ is the mean of that distribution, then the mean of the sample $x$ is a sufficient statistic for $\theta$. The generative process can be rewritten to generate the sample mean $s$ given $\theta$ and then generate the variation around that sample mean independently from $\theta$, and $s$ contains all the information relevant to estimating $\theta$.
%
In the autoregressive setting, we care about \textit{predictive sufficiency} \citep{Bernardo:1319894}. A statistic $s(x_{1:n})$ is predictive sufficient for the sequence $x_{1:n}$ if 
\begin{align}
p(x_{n+1}|x_{1:n}) = p(x_{n+1}|s(x_{1:n})).
\end{align}
Crucially, if a model performs autoregressive modeling perfectly, it learns a predictive sufficient statistic. 
%Our three cases thus correspond to settings where  

In the remainder of this section we describe two cases where predictive sufficient statistics can be easily identified and could plausibly be represented by a neural network (Figure \ref{fig:graphical-model}). These cases correspond to common applications of machine learning models as well as classic models for text such as hidden Markov models \citep{jurafsky2008speech}. First, when $x_{1:n}$ are independently sampled conditioned on an unknown parameter, $\phi_n$ needs only represent the sufficient statistic of this sequence. Second, when $x_{1:n}$ are generated by a state space model (in the discrete case, a hidden Markov model), $\phi_n$ needs only represent the posterior distribution over the next latent state given $x_{1:n}$. 
% Finally, when $x_{1:n}$ are sampled independently from one of a discrete set of latent distributions, $\phi_n$ needs only represent the posterior distribution over hypotheses about that distribution given $x_{1:n}$. 
In each case we explain how $p(x_{n+1}|x_{1:n})$ factorizes to make it possible for $x_{1:n}$ to be summarized by some $\phi_n$ and identify the form of the corresponding $g(\phi_n)$.

\subsection{Case 1: Exchangeable models}

Predictive sufficiency is straightforward to establish in {\em exchangeable} models, where the probability of a sequence remains the same under permutation of the order of its elements. That is, a sequence is exchangeable if $p(x_{1:N}) = p(x_{\pi(1:N)})$ for some permutation $\pi$. Any exchangeable model can be re-expressed in terms of the $x_i$ being sampled independently and identically distributed according to a latent distribution $p(x|\theta)$ parameterized by $\theta$, with $p(x_{1:N}) = \int_\theta \prod_i p(x_i|\theta)p(\theta) \, d\theta$ 
\citep{gelmanbda04}. This idea leads to the following proposition:

\textit{Proposition.} \label{sec:prop} Given an exchangeable sequence $x_{1:N}$ where each $x_i$ is of dimension $d_x$, and given functions $f: \mathbb{R}^{nd_x} \mapsto \mathbb{R}^{d_m}$, $g: \mathbb{R}^{d_m} \mapsto \mathbb{R}^{d_x}$ such that, for each $1 \leq n \leq N$, $g \circ f(x_{1:n}) = p(x_{n+1}|x_{1:n}) \; \forall x_{n+1}$, $f(x_{1:n})$ is a sufficient statistic for $x_{1:n}$. 

In other words, if we have a perfect autoregressive predictor that is composable into $g \circ f$, the output of $f$ is a sufficient statistic for its sequence input.
\begin{proof}
The result follows from the fact that for exchangeable sequences, general sufficiency is equivalent to predictive sufficiency \citep{Bernardo:1319894}.
Because $p(x_{n+1}|x_{1:n}) = g(f(x_{1:n})) \; \forall n$, $f(x_{1:n})$ is a predictive sufficient statistic for the sequence, and it is also a sufficient statistic.
\end{proof} 
% Finally, if the sequence $x_1, x_2,...$ is exchangeable, then we obtain an independence statement that motivates the choice of $s(x_{1:n})$ to uncover latent variables. The statement is, for a sufficient statistic $s(x_{1:n})$ and under exchangeability, $p(x_1,...,x_n|\theta, s)$ is independent of $\theta$. $\theta$ is the random variable (which is guaranteed to exist for exchangeable data) conditioned on which $x_{1:n+1}$ are i.i.d.
The resulting sufficient statistic also fully specifies the posterior on the parameters of the generating distribution, $p(\theta|x_{1:n})$. Sufficient statistics are easily identified for a wide range of distributions, including all exponential family distributions \citep{Bernardo:1319894}, and are easy to represent.
Since the LM's predictive distribution decomposes into the form above, this result gives us strong predictions about the contents of embeddings for models trained on exchangeable data. 

\subsubsection{Case 1.1: Discrete hypothesis spaces} 

In a more specific version of Case 1, assume each $x_i$ is generated independently from some unknown generative model, and let $\mathcal{H}$ denote the discrete set of hypotheses $h$ about the identity of this model. In this case, $x_{1:n}$ are exchangeable, but any sufficient statistics might be difficult to identify. The posterior predictive distribution can be written as
\begin{align*}
     p(x_{n+1}|x_{1:n}) = \sum_{h \in \mathcal{H}} p(x_{n+1}|h)p(h|x_{1:n}).
\end{align*}
By an argument similar to Case 2, $p(h|x_{1:n})$ is a predictive sufficient statistic in this model and an embedding thus need only capture this posterior distribution.

\subsubsection{Case 1.2: Topic Models} 
\label{sec:3-topic-models}

Latent Dirichlet Allocation (LDA; \citet{lda}) is an exchangeable generative model that is widely used for modelling the topic structure of documents. A document is generated from a mixture of $K$ topics. Each topic is a distribution over the vocabulary; e.g., a topic corresponding to geology might assign high probability to words such as \textit{mineral} or \textit{sedimentary}. We detail the full generative process in Appendix \ref{sec:appendix-models}.

After LDA is trained on a corpus, the inferred quantities can be used to explore the corpus. We denote by $\theta_i$ as the latent variable that stands for each document's underlying topic mixture. 
%An inferred latent variable commonly used to visualize and understand a topic model is $\beta_k$, which is a distribution over the vocabulary in topic $k$. 
%As an example, Figure \ref{fig:lda-topic} shows words with high values of $\beta_k$ for ten inferred topics.
Because topic mixture $\theta$ and word distribution $\beta$ form the sufficient statistic, our theory suggests that autoregressive LMs should implicitly encode the topic structure of a document. 

\subsection{Case 2: Latent state models}

%\leo{The below equation also applies to Case 1: Exchangeable Models, but the Proposition seems to make this equation redundant for exchangeable cases? If so, I wonder if it would still be worth writing this equation for Case 1 (I currently commented that out).}

In a latent state model, each $x_i$ is generated based on a latent variable $z_i$. These $z_i$ are interdependent, with $z_i$ being generated from a distribution conditioned on $x_{i-1}$. Common latent state models include Kalman filters (where $x_i$ and $z_i$ are continuous and the emission and transition functions, $p(x_i|z_i)$ and $p(z_i|z_{i-1})$, are linear-Gaussian) \citep{kalman1960new} and hidden Markov models (where the $z_i$ are discrete) \citep{baum1966statistical}. % TLG: give references for these
In a latent state model, the posterior predictive distribution is
\begin{align}
    \hspace{-2mm} p(&x_{n+1}|x_{1:n}) = \int p(x_{n+1}|z_{n+1})p(z_{n+1}|x_{1:n})\, dz_{n+1}. \label{eq:latentstate}
\end{align}
In this case, the posterior over the next latent state $p(z_{n+1}|x_{1:n})$ captures all of the information in $x_{1:n}$ relevant to predicting $x_{n+1}$, rendering $x_{n+1}$ independent of $x_{1:n}$ when conditioned on this statistic. If we were to drop conditional independence assumptions in a latent state model, one would need $p(x_{n+1}|z_{n+1}, x_{1:n})$ instead of $p(x_{n+1}|z_{n+1})$ for the equality. Thus, $p(z_{n+1}|x_{1:n})$ is a predictive sufficient statistic in this model. However, there is another predictive sufficient statistic. We can write
\begin{align}
    \hspace{-2mm} p(&x_{n+1}|x_{1:n}) = \int p(x_{n+1}|z_{n})p(z_{n}|x_{1:n})\, dz_{n}, \label{eq:latentstate2}
\end{align}
showing $p(z_{n}|x_{1:n})$ to also be a predictive sufficient statistic. Which of these will be favored will depend on how easily the relevant integrals can be approximated. 
%In other words, if an embedding is chosen to be a sequence representation learned by the autoregressive predictor (consider the last hidden state of the transformer), our analysis predicts that $p(z_{n+1}|x_{1:n})$ should be expected from this embedding.

More formally, we consider an embedding a representation $\phi_n$ such that a fixed operator $g(\phi_n)$ can produce $p(x_{n+1}|x_{1:n})$. The  distributions $p(z_{n+1}|x_{1:n})$ and $p(z_{n}|x_{1:n})$ satisfy this characterization, with Equations \ref{eq:latentstate} and \ref{eq:latentstate2} showing that the relevant operator is the integral of $p(x_{n+1}| z_{n+1})$ over $z_{n+1}$ or $p(x_{n+1}| z_{n})$ over $z_{n}$ and $z_{n+1}$. That operator can be easily approximated linearly and hence by a single layer of a neural network. The embedding $\phi_n$ thus needs only represent $p(z_{n+1}|x_{1:n})$ or $p(z_{n}|x_{1:n})$, depending on the relative ease of approximating the relevant integrals in a specific autoregressive predictor.

\subsection{Probing embeddings to recover predictive sufficient statistics}

These cases specify information that should be encoded in neural network embeddings. This sets up the second element of our approach, which is building probes to check this hypothesis. We focus on transformers \citep{vaswani2017attention}, since they are widely used as language models. 
We denote the decoding target for each sequence $x_{1:n}$ by a vector $t_n$. Given a trained transformer, we decode $t_n$ by training a second model (a probe) to predict the target $t_n$ from the embedding $\phi_n$ of the corresponding document $x_{1:n}$. The embedding is defined to be the last layer embedding, which is what researchers and practitioners typically use as a document representation.
The probe $g$ maps from the sequence embedding to the target. To ensure that the relevant statistical information is contained in the LM, not in the probe, we keep the probe simple by defining it as a linear layer with softmax activations: $g(\phi_n) = \text{Softmax}(\text{Linear}(\phi_n))$. In each case, the probe uses the last-layer, last-token embedding as its input, unless otherwise specified.

% \michael{here's i'm curious how performance degrades (or doesn't) as a function of the dimensionality of the latent hypothesis space}
% \michael{i'm curious how probe recovery results depend on the hypothesis space: discrete vs continuous}
% \subsection{Idea}

% We can use the sufficient statistic to clarify what is meant by `contains'. We can instead ask, whether $\phi(x_{1:n})$ is a sufficient statistic for $\theta$ or $z_{n+1}$. That is, whether the conditional distribution of $x_{n+1}$ given $\phi(x_{1:n})$ does not depend on $\theta$ or $z_{n+1}$. Mathematically, it is equivalent to asking whether the following equality holds: $I(\theta; \phi(x_{1:n})) = I(\theta; x_{n+1})$.

\section{Empirical analysis}
\label{sec:empirical}

We have identified cases where a predictive sufficient statistic is expected to be encoded by an autoregressive model. In this section, we use probing analyses on transformers to empirically validate this hypothesis. Ablation studies on memorization, parsimonious properties of embeddings, and decoding alternative quantities further support this idea. We use the Adam optimizer \citep{kingma2014adam} in all experiments. Code is available at \href{https://github.com/zhang-liyi/llm-embeddings}{github.com/zhang-liyi/llm-embeddings}.

\subsection{Simple exchangeable models}

\subsubsection{Generative model}

\paragraph{Bayesian conjugate models} 

We start with three data generating distributions: Gaussian-Gamma, Beta-Bernoulli, and Gamma-Exponential models. 

For $M$ Gaussian-Gamma sequences $\{x^{(k)}\}_{1\leq k\leq M}$ with $N$ tokens each, each sequence is i.i.d. generated by a mean and precision parameter sampled from a prior. 
%Across the $M$ sequences, $M$ mean and precision parameters are sampled to generate these sequences. 
For sequence ${x}^{(k)}$, the generative process is,
\vspace{-2mm}
\begin{align*}
    \tau_k &\sim \textsc{Gamma}(\alpha_0, \beta_0) \\
    \mu_k | \tau_k &\sim \mathcal{N}(\mu_0, (\lambda_0\tau_k)^{-1}) \\
    x^{(k)}_{n} | \mu_k,\tau_k &\sim \mathcal{N}(\mu_k, \tau_k^{-1}) \; \text{for} \; n \in \{1,...,N\},
\end{align*}
where $\alpha_0, \beta_0, \mu_0, \lambda_0$ are fixed hyperparameters. The full generative processes for the other two models are given in Appendix \ref{sec:appendix-models}. 

The Gaussian-Gamma model generates from a Gaussian distribution with two unknown parameters, mean $\mu$ and precision $\tau$. The predictive distribution for the next token $x_{n+1}$ is
\begin{align*}
    p(x_{n+1}|x_{1:n}) = \int p(x_{n+1}|\mu, \tau)p(\mu,\tau|x_{1:n})d(\mu,\tau).
\end{align*}
%In predictive predictive distributions, we drop $x$'s superscript $k$, and $x_{n+1}$ denotes the $n+1$-th token in sequence $x$. 
The optimal Bayesian agent uses the same prior distributions as the Gaussian and Gamma that generate $\mu$ and $\tau$. It will analytically infer the ground truth posterior $p(\mu,\tau|x_{1:n})$ for any stream of data it sees, and use this posterior for predicting next tokens. A Bayesian agent can also use other suitable priors and converge to the optimal posterior. To be consistent with other exchangeable conjugate models, we denote $\theta = (\mu, \tau)$ to indicate latent variables whose posterior distribution is predictive sufficient.

\paragraph{Discrete hypothesis spaces}

Each sequence in this dataset consists of two-dimensional points uniformly sampled from a rectangular region in 2D space \citep[cf.][]{tenenbaum1998bayesian}. The hypothesis space $\mathcal{H}$ is the set of all rectangles whose corner points are pairs of integers in $\{0, 1, 2, ..., 7\}$. The generative process uniformly samples a rectangle $h_{\text{rect}}$ from the set of all rectangles $\mathcal{H}$. Then, each token in a sequence is sampled uniformly from the region defined by $h_{\text{rect}}$. 

% For each sequence ${x}$, the generative process is,
% \begin{align*}
%     h_{\text{rect}} &\sim \mathcal{H}, \\
%     x_i &\sim \textsc{Uniform}(h_{\text{rect}}) \; \forall i.
% \end{align*}
% \michael{why not just }
% \michael{
% I'm wondering if there should be a negative control in the probing experiments. Like, making sure there are things that you shouldn't be able to decode, can't be decoded
% }
% \michael{I think you do this for HMM Experiments}
% \michael{are you training a separate probe for the OOD probe recovery results or a different one?
% }
% \leo{I'm training a separate probe. Perhaps a same probe would be interesting to try and might tighten the connection further?}
% \michael{are higher moments decoded with higher fidelity later than lower order moments}

\begin{figure}[t]
    \centering
    \begin{subfigure}{0.305\textwidth}
        \centering
        \includegraphics[width=\textwidth]{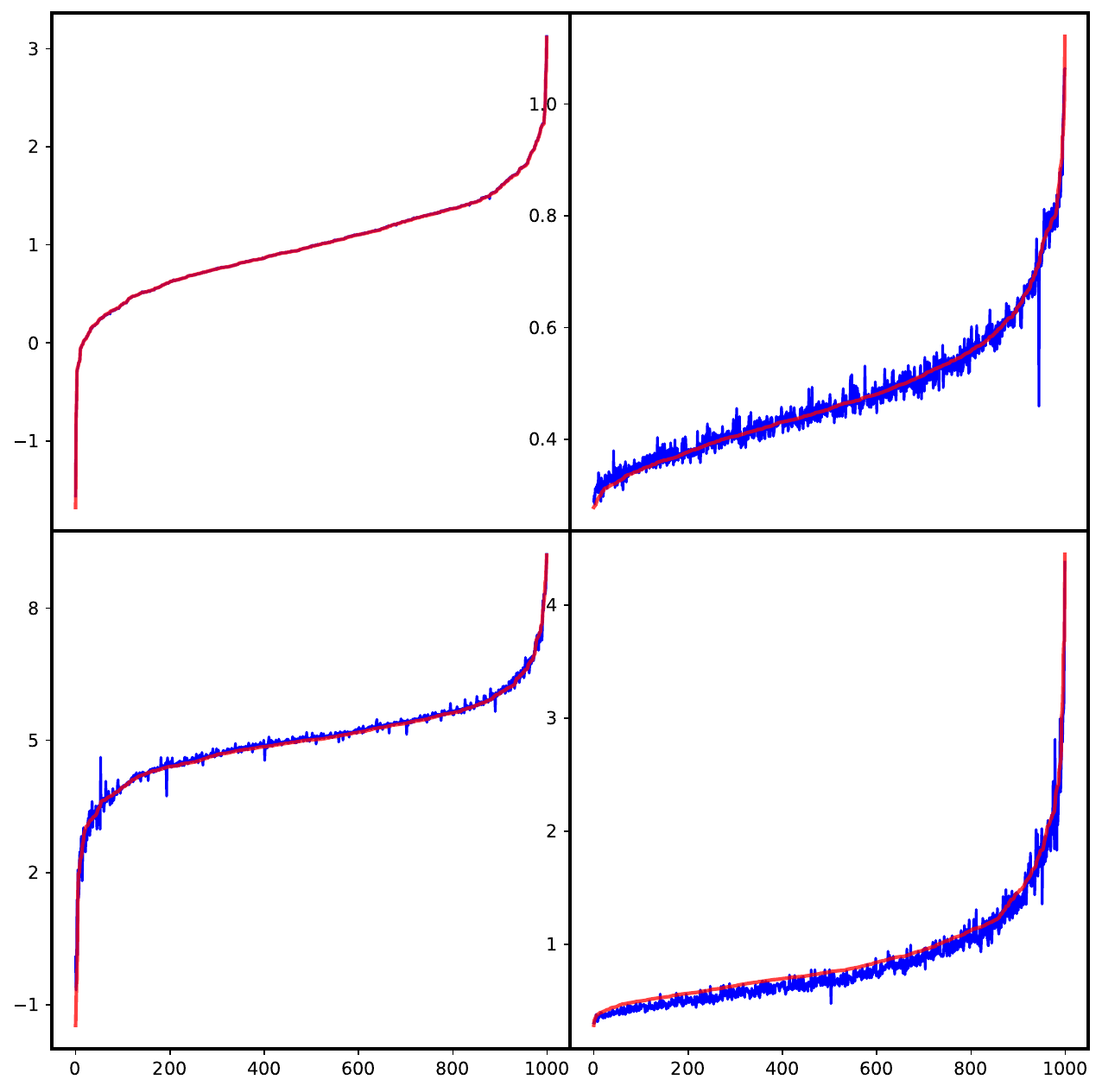}
        \caption{Gaussian-Gamma.}
    \end{subfigure}
    \begin{subfigure}{0.1615\textwidth}
        \centering
        \includegraphics[width=\textwidth]{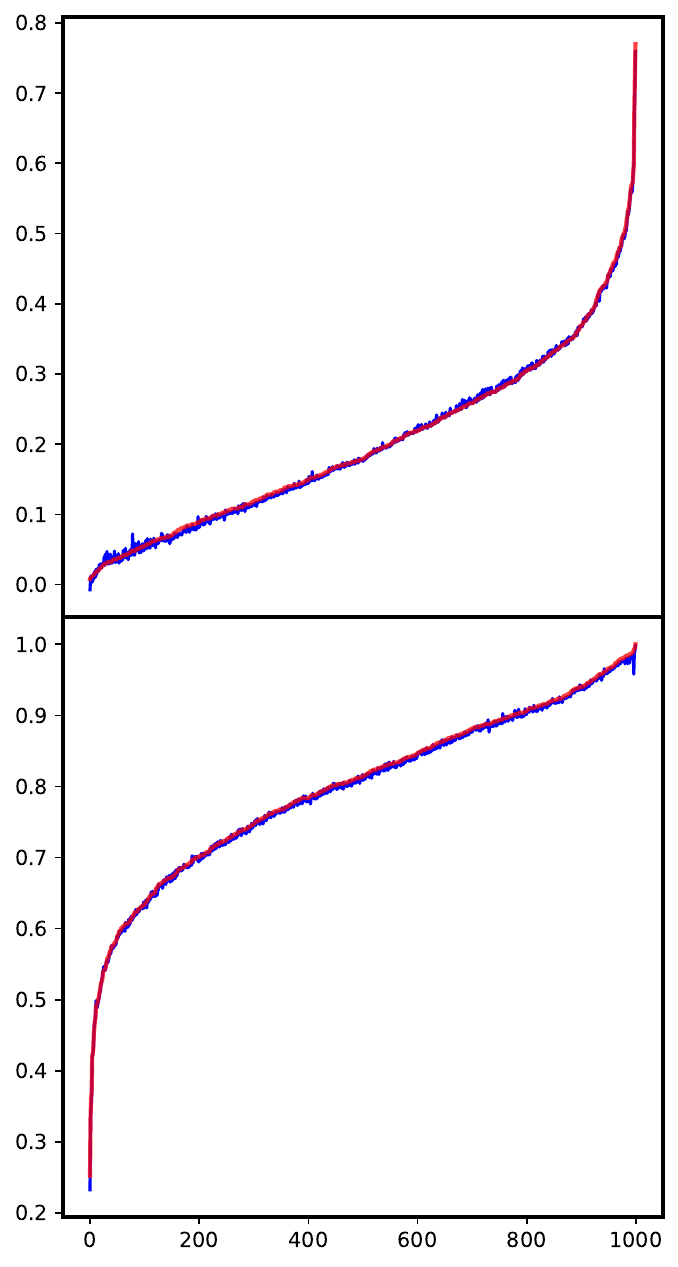}
        \caption{Beta-Bernoulli.}
    \end{subfigure}
    \begin{subfigure}{0.165\textwidth}
        \centering
        \includegraphics[width=\textwidth]{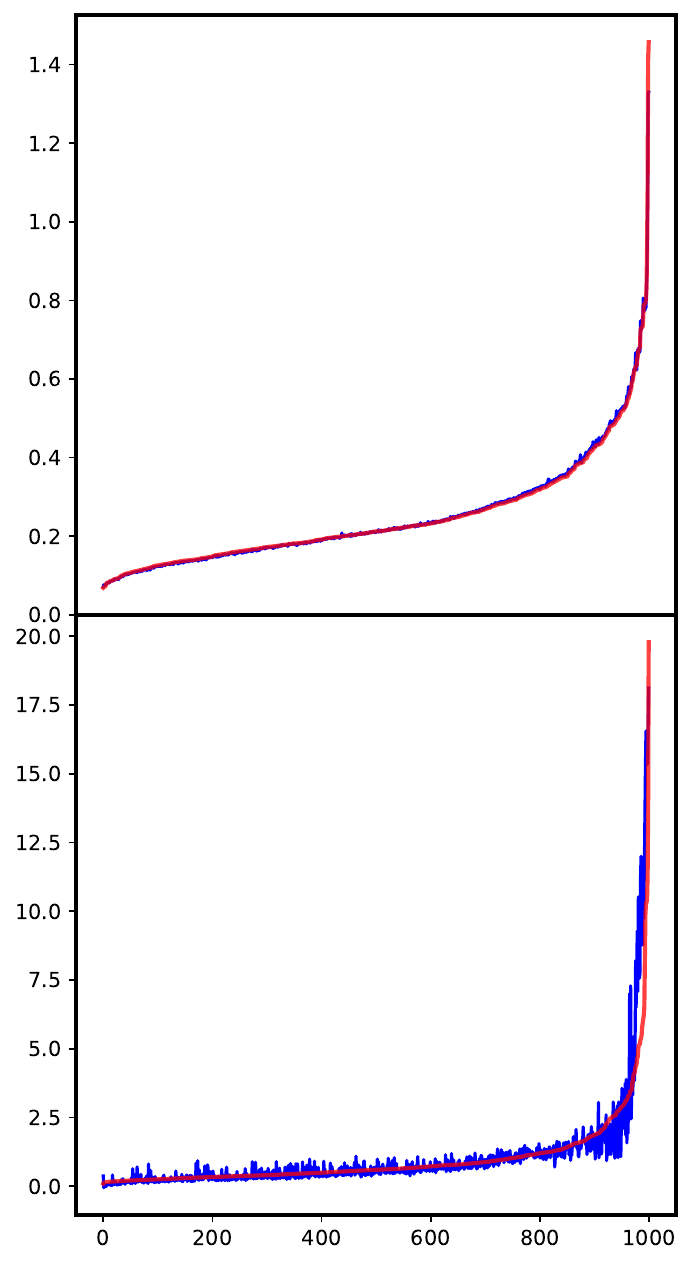}
        \caption{Gamma-Exp.}
    \end{subfigure}
    \caption{Probe recovery of transformer-learned sufficient statistic (blue) and ground truth sufficient statistic (red) on the y-axis, across 1000 test datapoints on the x-axis. In the plot above, the datapoints are sorted based on their ground truth sufficient statistic. The first row shows parameters probed in the non-OOD case (from left to right: Gaussian mean $\mu$, Gaussian precision $\tau$, Bernoulli mean, and Exponential mean). The second row shows the corresponding information in the OOD case.
    % \michael{if you have time to do runs with more conjugate hyperparameters, might be better to report this as a table with aggregated error across the 1000 testponts?}
    }
    \label{fig:conjugate-ss}
\end{figure}

% \begin{figure}[t]
%     \centering
%     \begin{subfigure}[b]{0.23\textwidth}
%         \centering
%         \includegraphics[width=0.8\textwidth]{figs/grid_integer.pdf}
%         \caption{A hypothesis space with equal-width rectangles.}
%         \label{fig:grid-integer}
%     \end{subfigure}
%     %\hspace{1cm}
%     \begin{subfigure}[b]{0.24\textwidth}
%         \centering
%         \includegraphics[width=0.8\textwidth]{figs/grid_decimal.pdf}
%         \caption{A hypothesis space with unequal-width rectangles.}
%         \label{fig:grid-decimal}
%     \end{subfigure}
%     \hfill
%     \caption{Two discrete hypothesis spaces $\mathcal{H}$ used in our experiments. Any continuous rectangle  contained within the axes (e.g., the red or the orange rectangle) is a valid hypothesis $h \in \mathcal{H}$. This results in 784 distinct hypotheses in each hypothesis space. The data consist of a sequence of points sampled uniformly from the target rectangle.}
%     \label{fig:grid}
% \end{figure}

% \begin{subfigure}[b]{0.178\textwidth}
    %     \centering
        
    %     \caption{A hypothesis space with unequal-width rectangles.}
    %     \label{fig:grid-decimal}
    % \end{subfigure}
    % \begin{subfigure}[b]{0.265\textwidth}
    %     \centering
    %     \includegraphics[width=\textwidth, trim={0 0.7cm 0 0},clip]{figs/pca_mean.pdf}
    %     \caption{Colored by mean of the generating rectangle.}
    %     \label{fig:pca-mean}
    % \end{subfigure}
    % \hfill

\begin{figure}[t]
    \centering
    \begin{subfigure}[b]{0.4\textwidth}
        \centering
        \includegraphics[width=0.475\textwidth]{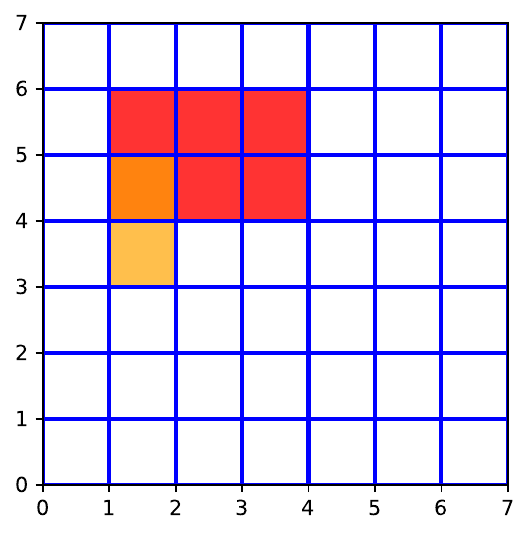}
        \includegraphics[width=0.505\textwidth]{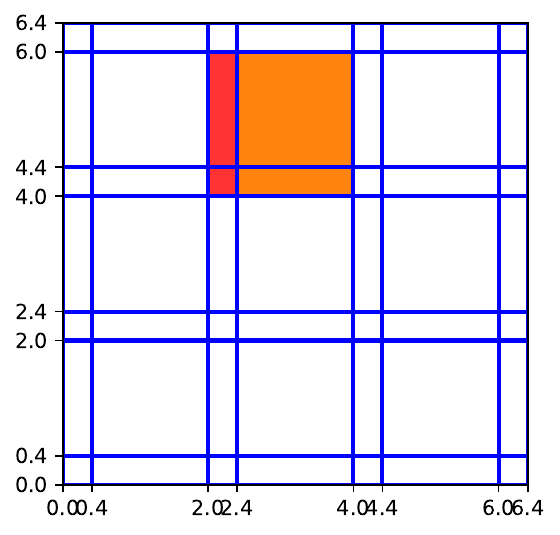}
        \caption{A hypothesis space with equal-width rectangles (left), and unequal widths (right).}
        \label{fig:grid}
    \end{subfigure}
    \hfill
    \begin{subfigure}[b]{0.265\textwidth}
        \centering
        \includegraphics[width=\textwidth, trim={0 0.7cm 0 0},clip]{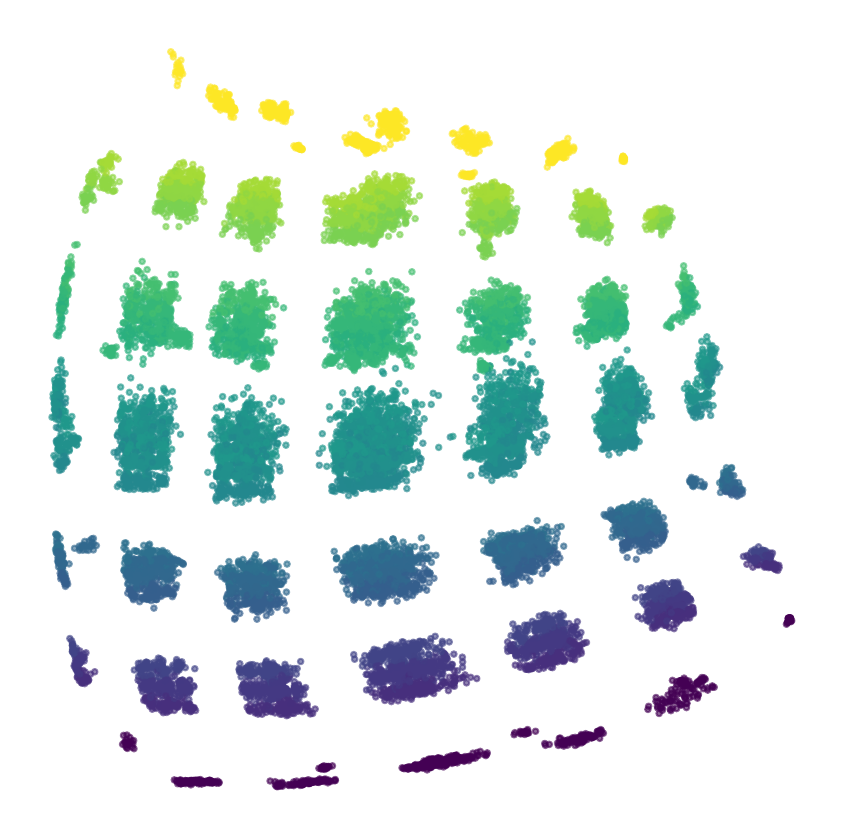}
        \caption{Colored by mean along y-axis of the generating rectangle.}
        \label{fig:pca-mean-y}
    \end{subfigure}
    \hfill
    \begin{subfigure}[b]{0.28\textwidth}
        \centering
        \includegraphics[width=\textwidth, trim={0 0.7cm 0 0},clip]{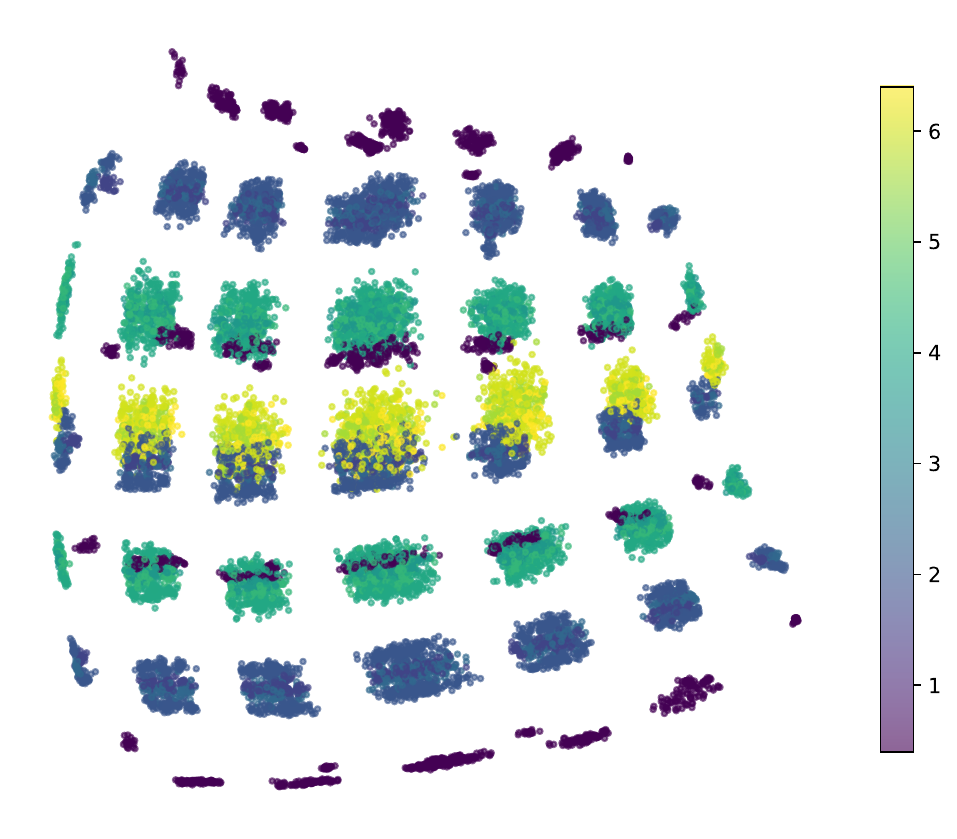}
        \caption{Colored by spread along y-axis of the generating rectangle.}
        \label{fig:pca-spread-y}
    \end{subfigure}
    %\vspace{-2mm}
    \caption{(a): Two discrete hypothesis spaces $\mathcal{H}$ used in experiments. Any continuous rectangle  contained within the axes (e.g., the red or the orange rectangle) is a valid hypothesis $h \in \mathcal{H}$. The data consist of a sequence of points sampled uniformly from the target rectangle. (b) and (c): Two-dimensional representation of embeddings of all validation datapoints (the setup is unequal width and sample size $=50$). The two subfigures show the same embeddings, colored by properties of the true generating rectangles.}
    \label{fig:pca}
\end{figure}

\begin{table}[t!]
    \centering
    \caption{Probing results on discrete hypothesis spaces. In general, the probe achieves strong performance in recovering a 784-length vector, and performance increases as task difficulty decreases. Average and standard error across 10 random seeds are reported.}
    \vspace{-2mm}
    \resizebox{0.55\columnwidth}{!}{
    \begin{tabular}{ccccc}
    \toprule 
    & Equal Width & & Unequal Width & \\ \cmidrule(lr){2-3} \cmidrule(lr){4-5}
    Sample Size & Accuracy $\uparrow$ & Squared Loss $\downarrow$ & Accuracy $\uparrow$ & Squared Loss $\downarrow$ \\
    \midrule  
    20 & $87.3 \pm 0.4\%$ & $0.173 \pm 0.005$ & $66 \pm 0.5\%$ & $0.29 \pm 0.004$ \\
    50 & $99.5 \pm 0.1\%$ & $0.008 \pm 0.001$ & $88.5 \pm 0.3\%$& $0.159 \pm 0.003$\\
    \bottomrule
    \end{tabular}}
    \label{tab:hypo}
\end{table}

\subsubsection{Probing experiments}

\paragraph{Implementation details} For each dataset, the out-of-distribution (OOD) hyperparameters are chosen such that OOD data are centered on a different mean and has a different spread. The exact choices are detailed in Appendix \ref{sec:appendix-implementation}, along with hyperparameter sweep in Appendix \ref{sec:appendix_results}. 

In discrete hypothesis spaces, we use rectangles with seven unit blocks on each side, resulting in a size-784 hypothesis space, i.e., 784 possible rectangles from which sequences are drawn. The unit lengths are either 1, or 0.4 and 0.16 alternating (Figure \ref{fig:grid}).

\paragraph{Experimental results}
We hypothesize that a transformer trained on these datasets should come to represent the sufficient statistics of the corresponding distribution. 
% We test if the true posterior $p(\theta|x_{1:N})$ can be uncovered from this transformer. 

\textit{Sufficient statistic.} Results for the three Bayesian conjugate models are shown in Figure \ref{fig:conjugate-ss}, where the probe successfully decodes the sufficient statistics $\theta$. For hypothesis spaces, we probe the distribution $p(h|x_{1:n})$, which is a length-784 simplex vector. Results are shown in Table \ref{tab:hypo}. In general, the true hypothesis can be found with high accuracy, even though the number of classes is high. 

Figure~\ref{fig:pca} visualizes how embeddings represent each hypothesis $h$ through Principal Components Analysis (PCA) that reduces the 128-dim embeddings to 2D, and the points are colored by properties of each rectangle.
Embeddings are clustered into different regions based on the geometry of the true generating rectangle (Figure \ref{fig:pca-mean-y} and \ref{fig:pca-spread-y}), where, in \ref{fig:pca-mean-y} for instance, the top yellow clusters correspond to the upper-most rectangles, and dark blue clusters correspond to the lower-most rectangles. Additionally, the embeddings encode the distance between corners of the generating rectangle (Figure \ref{fig:pca-spread-y}). We also find the same color pattern (but rotated 90 degrees) when the horizontal instead of vertical axis is used to color the embeddings.

\begin{figure*}[t]
    \centering
    \includegraphics[width=1\textwidth]{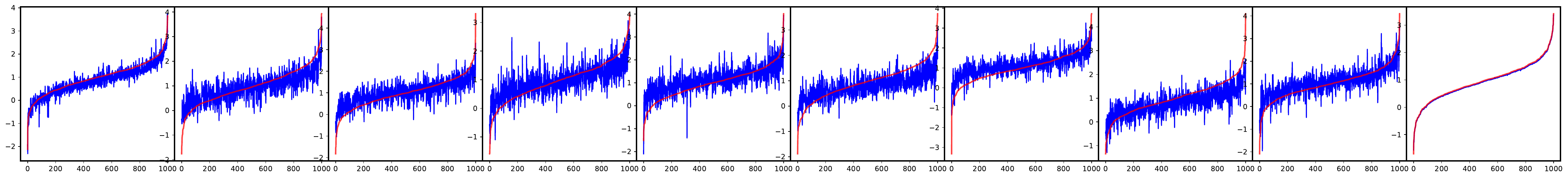}
    \vspace{-5mm}
    \caption{Probing over the first 10 tokens themselves using the 10th token embedding of the transformer. Aside from perfectly encoding the 10th token, this embedding does not show memorization over the other 9 tokens as suggested by the noise in probe recovery.}
    \label{fig:token-memorization}
\end{figure*}

\textit{Moments of the posterior distribution.} A consequence of knowing the sufficient statistic is finding the true posterior $p(\theta|x_{1:N})$, and we show that the moments of this posterior can be decoded from the transformer embedding (see Figure \ref{fig:gaussian-gamma-moment} and \ref{fig:conjugate-moment2} in the Appendix). The moments are functions of the mean and variance of the stream of observed data. In these conjugate models, it might be unsurprising for a transformer to encode the mean of the sequence it processes, because the optimal strategy for its loss function is to always predict the mean of the sequence that it sees so far. However, encoding the variance would not be directly related to this strategy and would support the argument that it infers sufficient statistics. 

\textit{Out-of-distribution simulations.} The analytical nature of the Bayesian predictor means that it is robust to datasets generated far from the prior -- it would simply update its posterior based on the data. Thus, we probe the transformer on out-of-distribution (OOD) datasets that are generated from a distribution of the same form but with distinct hyperparameters. Figure \ref{fig:conjugate-ss}, as well as \ref{fig:gaussian-gamma-moment1} in the Appendix, suggest that the recovery of sufficient statistic and moments of the posterior distribution is generalizable to OOD cases, just as a Bayes-optimal agent would generalize. 
% \michael{it's probably worth adding a OOD section to the study design section and flagging it because reviewers will definitely look for it}
% \michael{also i think addding some details on the transformer architecture is important too}
% \michael{do you have checkpoints saved from training? if so could be interesting to assess probe recover at different checkpoints during training}
% \michael{are you planning to experiment with different $\alpha_0, \beta_0$ parameters for the OOD setting?  We could characterize the probe recovery performance as a function of the gap between the training hyperparameters and the OOD hyperparameters. For example, you coould vary the mean but not the spread and the spread but not the mean and check how probe recovery changes as a function of that.
% }

\paragraph{Memorizing the context or storing sufficient statistics of the context?} 

A possible confound behind the probe is that, instead of having a probe that uses the transformer-embedded sufficient statistics, the transformer may memorize a set of tokens, and the probe infers quantities of interest from these memorized inputs. We begin by making sure that the sufficient statistics over the first ten tokens can be decoded from the probe on the tenth token in the Gaussian-Gamma case (Figure \ref{fig:token10} in Appendix \ref{sec:appendix_results}). Then, we probed the token values themselves to look for memorization. Figure \ref{fig:token-memorization} suggests that memorization is generally absent. The 10th token embedding recovers the 10th token perfectly, but cannot recover the other 9 tokens. For the other 9 tokens, a correlation exists between probe results and true token values, but some correlation is expected because even a single token can reveal information about the generating distribution. However, the noise suggests that the model is finding sufficient statistics rather than memorizing.

\paragraph{Embeddings are parsimonious} 

Our analysis so far has focused on whether embeddings contain information that allows decoding  sufficient statistics. However, our hypothesis is stronger than this: since the sufficient statistics are enough to encode the relevant information from the data, a model need \textit{only} represent that information. To explore whether transformers construct such parsimonious representations, we fixed the models trained on different distributions above and used a multi-layer perceptron (MLP) to directly predict their last-layer, last-token embeddings using the sufficient statistics of the training sequences as input.

Results show that in most cases the sufficient statistics capture well over 50\% of variance in the embeddings (Table \ref{tab:parsimonious} in Appendix \ref{sec:appendix_results}). The results show that sufficient statistics capture a substantial amount of variance in transformer embeddings, establishing that the relationship between embeddings and sufficient statistics runs in both directions: we can decode sufficient statistics from embeddings, and we can predict embeddings from sufficient statistics.

\begin{table}[t] 
\centering 
\caption{Probing target quantities in HMM dataset using different transformer token embeddings. The target $p({z}_{n+1}|{x}_{1:n})$ is a simplex vector found by running the forward-backward algorithm, and the target $\hat{{z}}_{n+1}$ is a scalar standing for the most likely latent class found by Viterbi algorithm. Average and standard error across 10 random seeds are reported.} 
\vspace{-2mm}
\resizebox{0.65\columnwidth}{!}
{ \begin{tabular}{cccccc} 
\toprule & & $\delta=0.5$ & & $\delta=1$ & \\ 
\cmidrule(lr){3-4} \cmidrule(lr){5-6} Target Quantity & Embedding & Accuracy $\uparrow$ & Squared Loss $\downarrow$ & Accuracy $\uparrow$ & Squared Loss $\downarrow$ \\ 
\midrule 
$p({z}_{n+1}|{x}_{1:n})$ & ${x}_{1:n}$ & $90.8 \pm 1.7\%$ & $0.011 \pm 0.004$ & $90.4 \pm 1.9\%$ & $0.011 \pm 0.003$\\ 
$p({z}_{n+1}|{x}_{1:n+1})$ & ${x}_{1:n+1}$ & $86.5 \pm 1.3\%$ & $0.066 \pm 0.011$ & $82.2 \pm 1.8\%$ & $0.067 \pm 0.011$\\
\midrule 
$p({z}_{n+1}|{x}_{1:n})$ & ${x}_{1:n+1}$ & $66.6 \pm 6.6\%$ & $0.072 \pm 0.013$ & $65.9 \pm 6.8\%$ & $0.058 \pm 0.011$\\
$p({z}_{n+1}|{x}_{1:n+1})$ & ${x}_{1:n}$ & $53.2 \pm 3.5\%$ & $0.356 \pm 0.014$ & $50.8 \pm 3.6\%$ & $0.278 \pm 0.012$\\
$\hat{{z}}_{n+1}$ & ${x}_{1:n}$ & $59.8 \pm 3.8\%$ & / & $61.4 \pm 4.7\%$ & /\\ 
$\hat{{z}}_{n+1}$ & ${x}_{1:n+1}$ & $80.8 \pm 2.2\%$ & / & $77.5 \pm 2.8\%$ & /\\ 
\bottomrule 
\end{tabular}} 
\label{tab:hmm}
\end{table}

\subsection{Hidden Markov model}

\subsubsection{Generative model}

For an HMM generating data with $M$ sequences with $N$ tokens each, we formulate the generative process as
\vspace{-0.7mm}
\begin{align*}
    A_c &\sim \textsc{Dirichlet}_C(\gamma) \; \text{for} \; c \in \{1,...,C\} \\
    B_c &\sim \textsc{Dirichlet}_V(\delta) \; \text{for} \; c \in \{1,...,C\} \\
    {z}_0 &\sim \textsc{Categorical}({\pi}) \\
    {x}_{i} &\sim \textsc{Categorical}(B_{{z}_{i}}) \\
    {z}_{i+1} &\sim \textsc{Categorical}(A_{{z}_{i}}),
\end{align*}
where $C, V, {\pi}$ are initialized and denote, respectively, the number of classes, vocabulary size, and a list of probabilities on the number of classes to initialize the first latent state. $\gamma, \delta$ are scalar hyperparameters that are also initialized and fixed, and they represent the evenness of the samples from the Dirichlet distributions. $A$, $B$, and ${z}$, as a result, represent the transition matrix, the emission matrix, and the latent states, respectively.

We use the forward-backward algorithm \citep{rabiner89} to compute the posterior $p({z}_{n+1}|{x}_{1:n})$, and explore whether this distribution can be decoded from the transformer embedding on ${x}_{1:n}$. 

\subsubsection{Probing experiments}

%One of our targets is the most likely sequence of latent states given the whole observed sequence. This target is found by the Viterbi algorithm and is related to but distinct from the predictive sufficient statistic, $p(z_{n+1}|x_{1:n})$.

%\textit{Belief state inference.} 

%Specifically, the Viterbi algorithm gives a single most likely latent class ${z}_{n+1}$ given the whole sequence ${x}_{1:N}$. On the contrary, the transformer embedding that supposedly encodes ${z}_{n+1}$ only has information on ${x}_{1:n}$.

\paragraph{Implementation details}
We choose $C=4, V=64, \gamma=0.5$, and set ${\pi}$ to be uniform in our experiments. We also vary $\delta$ to control the level of difficulty: how distinct is one class from another.

\paragraph{Results}

Our theoretical treatment suggests that the transformer should encode the predictive sufficient statistic $p(z_{n+1}|x_{1:n})$ or $p(z_{n}|x_{1:n})$ . The latter is the target used in belief state inference with transformers \citep{shai2024transformersrepresentbeliefstate}.
However, there are other natural decoding targets that could be used. A simple target is the one-hot vector corresponding to the most likely hidden state $\hat{z}_{n+1}$. This is not a predictive sufficient statistic and does not encode the full information about the sequence relevant to future prediction. Our analysis thus suggests that it will thus provide a poorer match to the information contained in the embedding.

Table \ref{tab:hmm} suggests that the transformer encodes the predictive sufficient statistics. Furthermore, performance is better for our hypothesized measure (first and second row) than for related quantities (all other rows). Additionally, it is more difficult to recover the measure used in belief state inference (second row), although it is intuitively easier to learn: the target probability is over the same latent variable, but the emission of this latent variable is observed, unlike in our hypothesized measure where it is unobserved. This may be because if the embedding were to directly encode $p(z_{n}|x_{1:n})$, the transformer output layer after the embedding would additionally need to encode the transition matrix and additional integration computations, potentially making the task more challenging than encoding $p(z_{n+1}|x_{1:n})$.

% In particular, the transformer embeddings do not keep track of quantities irrelevant to prediction given the predictive sufficient statistics. Moving one token ahead, the embedding on the measure $p(z_{n+1}|x_{1:n})$ is eroded when the target is $x_{n+2}$

% Several quantities exhibit high volatility across random seeds, since the transition matrix has a greater influence on results when the tokens conditioned in the target posterior and the token used in the embedding mismatch.

\subsection{Topic Models}

We aim at recovering from language models the topic mixture $\theta_i$ that draws words in document $i$. 

\begin{table}[t]
\centering
\caption{Topic prediction by the autoregressive transformer (AT), \textsc{Bert}, LDA, and word-embedder (WE) on LDA-generated synthetic datasets, along with standard deviations across 3 random seeds. Hyperparameter $\alpha$ defines the dataset generation process, where a higher $\alpha$ means a more difficult task with underlying topics being more evenly distributed. AT and \textsc{Bert} have similar performance in the easiest setting, but AT performs well in harder settings where \textsc{Bert} performances worsen. End-to-end WE achieves stronger performance than language models, and LDA matches expectations by providing an upper bound in performance.}
\resizebox{0.5\columnwidth}{!}
{\begin{tabular}{ccccc} 
 \toprule
 $\alpha$ & Method & Accuracy $\uparrow$ & L2 loss $\downarrow$ & Tot. var. loss $\downarrow$\\
 \midrule
 & AT & $82.8\% \pm 0.5\%$ & $0.041 \pm 0.001$ & $0.141 \pm 0.001$ \\
 0.5 & \textsc{Bert} & $83.6\% \pm 1\%$ & $0.036 \pm 0.003$ & $0.131 \pm 0.005$\\
 & LDA & $87\% \pm 0.6\%$ & $0.029 \pm 0$ & $0.117 \pm 0.001$\\
 & WE & $85.8\% \pm 1.3\%$ & $0.03 \pm 0.001$ & $0.119 \pm 0.002$ \\
 \midrule
 & AT & $75.5\% \pm 0.8\%$ & $0.044 \pm 0.001$ & $0.144 \pm 0.001$ \\
 0.8 & \textsc{Bert} & $51.5\% \pm 1.7\%$ & $0.111 \pm 0.005$ & $0.233 \pm 0.011$\\
 & LDA & $82.6\% \pm 0.5\%$ & $0.036 \pm 0.001$ & $0.133 \pm 0.004$\\
 & WE & $80.9\% \pm 0.5\%$ & $0.029 \pm 0$ & $0.116 \pm 0.001$ \\
 \midrule
 & AT & $70.5\% \pm 1.6\%$ & $0.045 \pm 0.001$ & $0.146 \pm 0.003$ \\
 1 & \textsc{Bert} & $46.6\% \pm 3.3\%$ & $0.1 \pm 0.004$ & $0.222 \pm 0.006$\\
 & LDA & $79.6\% \pm 1.4\%$ & $0.045 \pm 0.004$ & $0.147 \pm 0.006$\\
 & WE & $79.4\% \pm 1\%$ & $0.027 \pm 0$ & $0.113 \pm 0.001$ \\
 \bottomrule
\end{tabular}}
\label{tab:toy-lm-lda}
\end{table}

\begin{figure*}[t]
    \centering
    \begin{subfigure}[b]{0.203\textwidth}
        \centering
        \includegraphics[width=\textwidth]{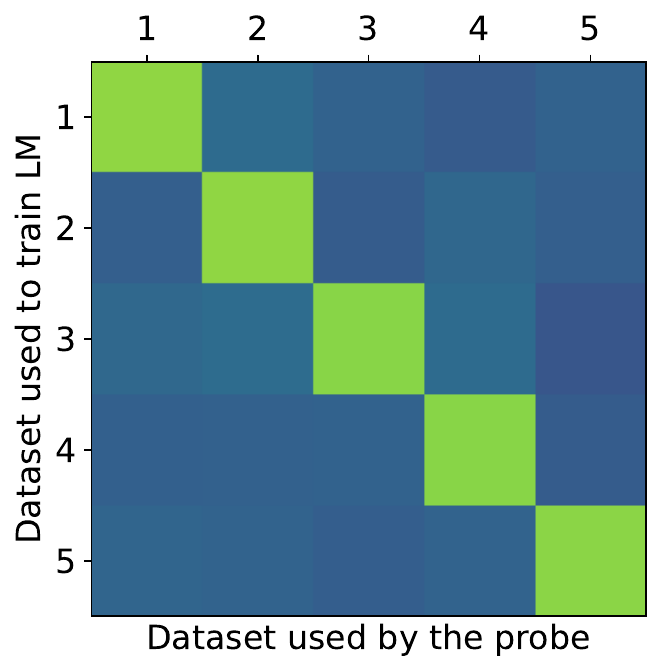}
        \caption{AT probe accuracies.}
        \label{fig:heatmap-lm}
    \end{subfigure}
    \begin{subfigure}[b]{0.243\textwidth}
        \centering
        \includegraphics[width=\textwidth]{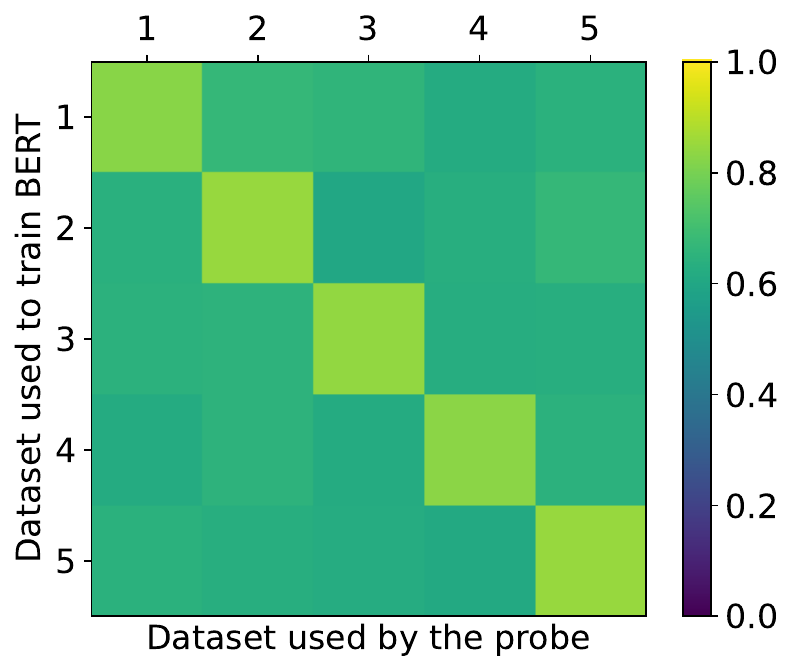}
        \caption{\textsc{Bert} probe accuracies.}
        \label{fig:heatmap-bert}
    \end{subfigure}
    \begin{subfigure}[b]{0.27\textwidth}
        \centering
        \includegraphics[width=1\textwidth]{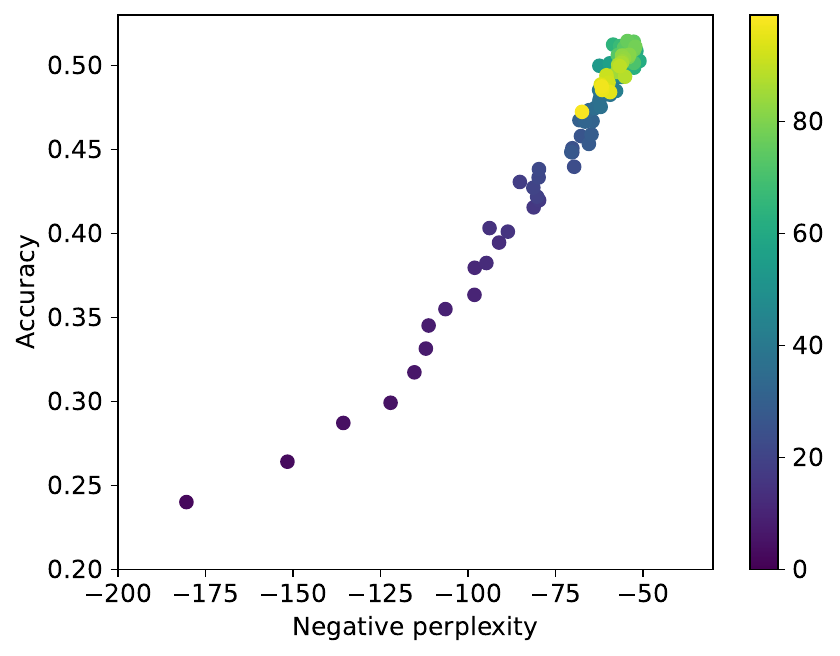}
        \caption{Accuracy vs. neg. perplexity}
        \label{fig:token-acc-perp}
    \end{subfigure}
    
    \caption{Figure \ref{fig:heatmap-lm} and \ref{fig:heatmap-bert}: control experiments showing AT (left) and \textsc{Bert}'s (middle) probe validation performance on synthetic data. For each AT and \textsc{Bert}, five models are trained and validated on five datasets, with each dataset generated by a distinct topic model. Colors show probe accuracy. A cell on row $i$ and column $j$ corresponds to model $i$ on dataset $j$, so the diagonal corresponds to a model on its own dataset. {\it For AT, performance is only strong on the dataset with the same generating topic model, suggesting that the underlying statistical model, not the probe taking different word embeddings, is responsible for performance -- a relationship that is also present for \textsc{Bert} but to a lesser degree.} Figure \ref{fig:token-acc-perp}: 20NG probe classification performance (accuracy) vs.\ negative perplexity measured at 100 different tokens. The dots are colored by the position percentile. {\it Probe performance increases with lower perplexity}.}
    \label{fig:heatmap-synthetic}
\end{figure*}

\subsubsection{Probing experiments}
\label{sec:4.1}

\paragraph{Experiment setup}
The dataset is bags-of-words generated by LDA. We set the vocabulary size $V=10^3$, number of topics $K=5$, and generated $N=10^4$ documents that are each 100 words long. 
% The steps made in this case study are,
% \begin{enumerate}
%     \item Draw bags of words \textbf{x} from topic model, $\textbf{x} \sim \text{LDA}(\alpha, \eta)$.
%     \item Train a language model on corpus \textbf{x}.
%     \item Train a linear classifier that takes the language model embedding for each document, and targets ground truth topic proportions $\theta$. 
% \end{enumerate}
% Unlike the experiments on natural corpora, the language models were initialized at random and trained on the dataset. Additionally, the topic probe was trained to target the ground truth topic mixtures rather than those inferred by LDA.

\paragraph{Models} We trained four models: an autoregressive transformer decoder (AT), \textsc{Bert}, LDA, and an end-to-end word embedder (WE). \textsc{Bert} uses a masked objective instead of the autoregressive one, and is implemented by a small version called \textsc{Bert-tiny} \citep{Turc2019WellReadSL}. 
LDA is used to establish an upper-bound for model performance. 
% Meanwhile, it remains a question whether high performance reveals that the model is really learning the underlying topic model, or is the topic probe mainly responsible for identifying topics. To test this idea, we use
We also include a model intended to provide an upper bound for embedding performance: a word embedder which is a matrix that maps from the vocabulary space to the AT / \textsc{Bert} embedding dimension and is end-to-end trained with the probe that predicts topics. 

For each model, the hidden sizes and final-layer embedding sizes are 128. AT has 4 decoder layers to match the size of \textsc{Bert-tiny}. More hyperparameters are detailed in Appendix \ref{sec:appendix-implementation}. 

\paragraph{Metrics} We use accuracy, L2 loss, and total variation loss to measure both classification performance and recovery of the topic mixture distribution. Accuracy is defined as how often the top topic predicted by the classifier's mixture agrees with the top topic from ground truth. The remaining loss measures apply to the whole topic mixture.

\begin{figure}
    \centering
    \includegraphics[width=0.7\linewidth]{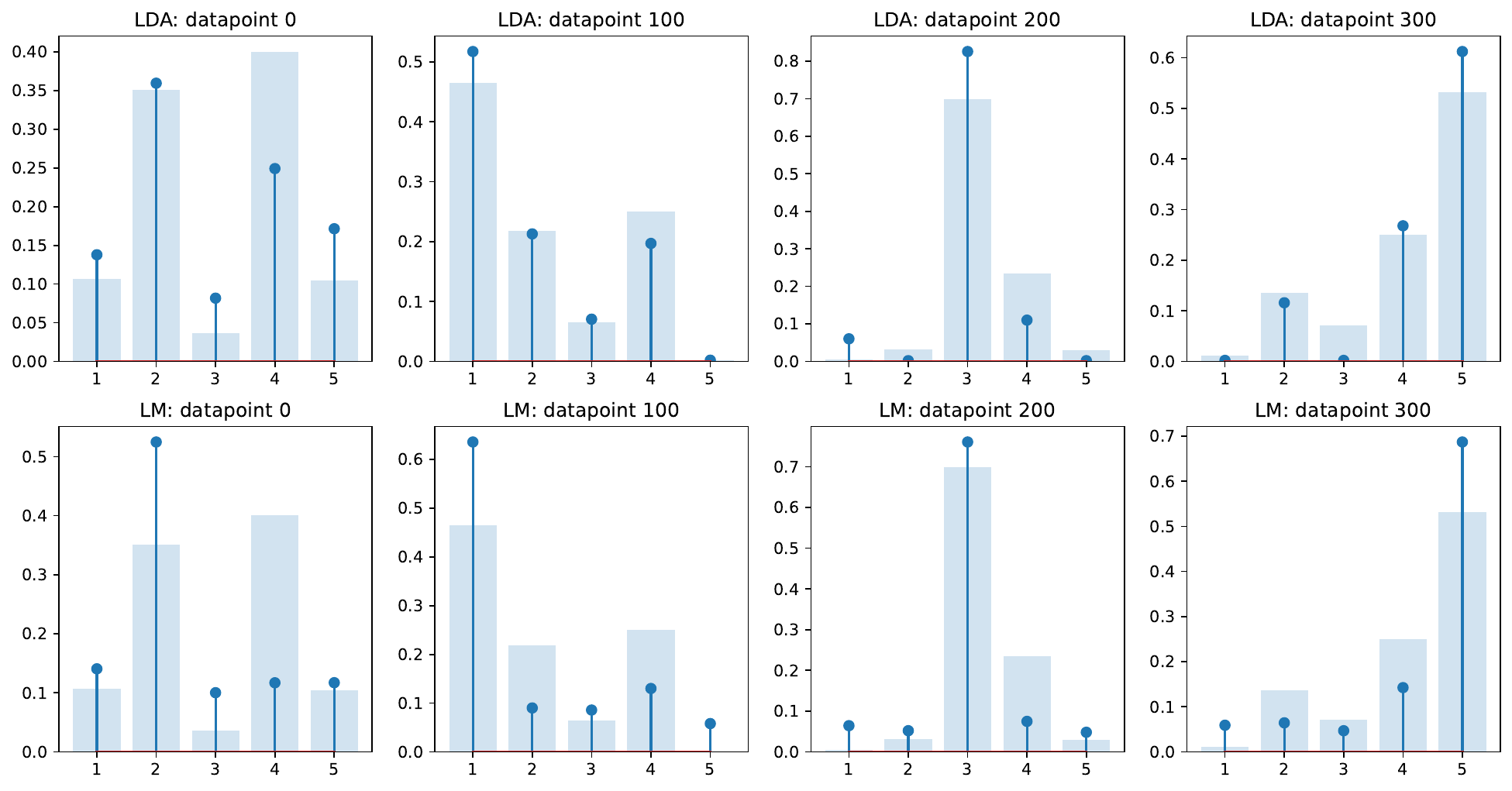}
    \caption{Distribution of synthetic data topics predicted by LDA (row 1) and AT classifier (row 2) for different validation datapoints. Predictions are stick plots, and ground truth is bar chart in the background. They both exhibit learning of topic mixtures by trying to match the distribution, in addition to top-1 agreement.}
    \label{fig:stickplots}
\end{figure}

\paragraph{Results} Figure \ref{fig:stickplots} shows five examples of ground truth, transformer-predicted, and LDA-predicted topic mixtures. Table \ref{tab:toy-lm-lda} shows that all models demonstrate success at recovering latent topics on at least the easiest setting (i.e., $\alpha=0.5$), being able to return both top-topic accuracy and the topic distribution spread. Between AT and \textsc{Bert}, the probe on AT is able to infer latent topic structures in more difficult tasks (i.e., $\alpha=0.8, 1$) whereas the probe on \textsc{Bert} shows deteriorating performance. LDA outperforms both AT and \textsc{Bert} as expected because it is specified exactly to learn a dataset generated by the other manually initialized LDA. However, the strong WE performance suggests that the probe can predict topics mainly from stand-alone words. This raises the question whether AT and \textsc{Bert} are learning an underlying statistical model, or are simply uniquely embedding each word and making the probe mainly responsible for topic recovery.

\subsubsection{Controlling for probe performance}

In this section, we conduct control experiments that suggest that language models learn an underlying statistical model, making them---rather than the topic probes taking word embeddings ---mainly responsible for successful topic recovery. 
If AT or \textsc{Bert} performs well just because it gives each word a unique embedding from which a trained probe suffices to recover the topic mixture, then a probe on top of the language model should additionally predict topic mixtures from a different underlying topic model than the one that generated the language model's training data. 

To assess this, we generate five datasets using five distinct topic models under the setting of $\alpha=0.5$. By definition, these topic models initialized five different sets of topics, or distributions over words $\{\beta_1,...,\beta_K\}_j, j\in\{1,2,3,4,5\}$. The transformer trained on set $j$ should encode the mixture of topics $\theta$ that corresponds to topics $\{\beta_1,...,\beta_K\}_j$.

One AT and one \textsc{Bert} is trained on each dataset. On each model, five probes are used to predict topics from each of the five datasets. Results are shown in Figure \ref{fig:heatmap-synthetic}. AT shows a strong distinction between predicting its own dataset versus  datasets from other topic models, whereas this distinction is present but weaker for \textsc{Bert}.
These results are evidence that AT and \textsc{Bert} indeed encode topic information because, if the topic information were instead constructed by the probe, the probe would work equally well on mismatched datasets as on matched ones. 

\subsection{From HMM-LDA to Natural Corpora}

In this section, we start with a synthetic dataset with a more realistic assumption than topic models --- that topics are an exchangeable component in a partially exchangeable sequence.
We generate a dataset based on both LDA and a hidden Markov model (HMM). Then, we move to evaluate the extent to which LLMs recover topic mixture two natural corpora, \textit{20Newsgroups} (20NG) and \textit{WikiText-103}. 

% \begin{table*}[h]
%     \centering
%     \caption{WikiText-13 accuracy for embeddings taken where the model outputs words that belong to a parts-of-speech category}
%     \resizebox{1.38\columnwidth}{!}{
%     \begin{tabular}{ccccc}
%     \toprule Model & Parameters & Noun & Verb & Adposition \\
%     \midrule
%        \textsc{Gpt-2}  & 124M & $69.5\% \pm 1.6\%$ & $70.2\% \pm 2.1\%$ & $71.7\% \pm 1\%$ \\
%        \textsc{Gpt-2-medium} & 355M & $69.2\% \pm 1.8\%$ & $69\% \pm 1.8\%$ & $72.1\% \pm 1.3\%$ \\
%        \textsc{Gpt-2-large} & 774M & $70.2\% \pm 1.1\%$ & $69.5\% \pm 1.7\%$ & $72.3\% \pm 0.9\%$ \\
%        \textsc{Llama 2} & 7B & $70.4\% \pm 1.9\%$ & $71.1\% \pm 2.2\%$ & $73.4\% \pm 1.8\%$ \\
%        \textsc{Llama 2-chat} & 7B & $71.2\% \pm 1.8\%$ & $72.1\% \pm 1.9\%$ & $76\% \pm 1.5\%$ \\
%     \midrule
%        \textsc{Bert} & 110M & $73.4\% \pm 1.2\%$ & $72.5\% \pm 1.3\%$ & $71.5\% \pm 1.3\%$ \\
%        \textsc{Bert-large} & 336M & $64.6\% \pm 1.8\%$ & $62.5\% \pm 1.9\%$ & $66.6\% \pm 1.3\%$ \\
%     \bottomrule
%     \end{tabular}}
%     \label{tab:20ng}
% \end{table*}

\begin{figure*}[t]
    \centering
    \includegraphics[width=13.2cm]{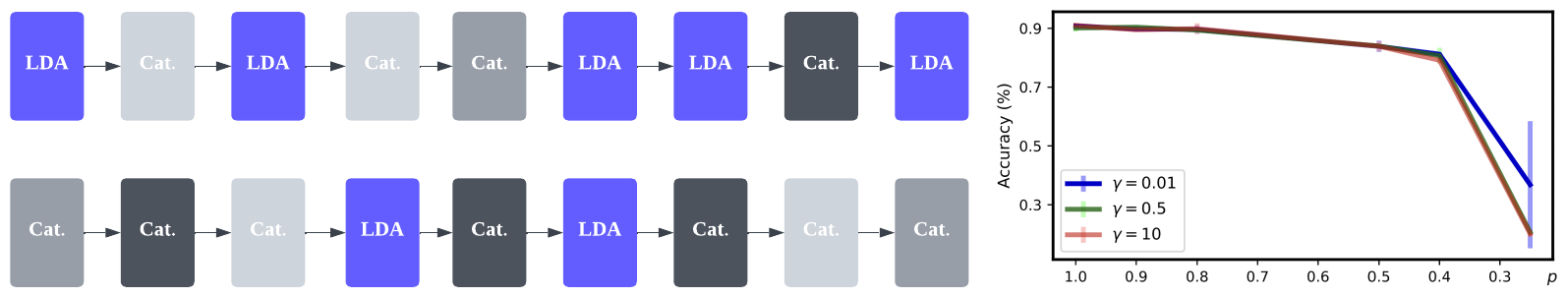}
    \caption{Left: Illustration of two sequences (upper: $p\approx 0.5$; lower: $p\approx 0.25$). Each box is a word, coming from either the LDA-generated semantic component (blue), or one of the three categorical distributions that indicate three syntactic classes (grey). Right: Probing accuracy across semantic proportion $p$ on HMM-LDA generated datasets. Lower $\gamma$ corresponds to more sparse transition matrices. \textit{The language model successfully encodes semantics if it occupies a considerable enough proportion.}}
    \label{fig:hmm-lda}
\end{figure*}

\begin{table*}[t]
    \centering
    \caption{20NG topic prediction performance based on different LLMs. \textit{Trained LLMs substantially outperform the null GPT-2 model, supporting the hypothesis that the training process encourages LLMs to implicitly develop topic models. Autoregressive models (the first five models) statistically significantly outperform non-autoregressive models (the next two).}}
    \resizebox{1\columnwidth}{!}{
    \begin{tabular}{cccccccc}
    \toprule 
    & & & $K=20$ & & & $K=100$ & \\ \cmidrule(lr){3-5} \cmidrule(lr){6-8} 
    Model & Parameters & Accuracy $\uparrow$ & L2 loss $\downarrow$ & Tot. var. loss $\downarrow$ & Accuracy $\uparrow$ & L2 loss $\downarrow$ & Tot. var. loss $\downarrow$ \\
    \midrule
       \textsc{Gpt-2}  & 124M & $61.4\% \pm 1.5\%$ & $0.106 \pm 0.002$ & $0.211 \pm 0.001$ & $42.3\% \pm 2.4\%$ & $0.097 \pm 0$ & $0.192 \pm 0.001$\\
       \textsc{Gpt-2-medium} & 355M & $62.6\% \pm 1.7\%$ & $0.104 \pm 0.002$ & $0.209 \pm 0.002$ & $42.9\% \pm 2.4\%$ & $0.096 \pm 0$ & $0.19 \pm 0.001$\\
       \textsc{Gpt-2-large} & 774M & $62.4\% \pm 1.8\%$ & $0.102 \pm 0.002$ & $0.208 \pm 0.002$ & $43.1\% \pm 2.3\%$ & $0.095 \pm 0.001$ & $0.189 \pm 0$\\
       \textsc{Llama 2} & 7B & $62.6\% \pm 1.7\%$ & $0.101 \pm 0.002$ & $0.206 \pm 0.002$ & $43.3\% \pm 2.4\%$ & $0.095 \pm 0.001$ & $0.189 \pm 0.001$\\
       \textsc{Llama 2-chat} & 7B & $62.9\% \pm 1.7\%$ & $0.102 \pm 0.002$ & $0.207 \pm 0.002$ & $43.2\% \pm 2.5\%$ & $0.095 \pm 0.001$ & $0.189 \pm 0$\\
    \midrule
       \textsc{Bert} & 110M & $56.3\% \pm 1.5\%$ & $0.113 \pm 0.003$ & $0.222 \pm 0.003$ & $38.6\% \pm 2.5\%$ & $0.1 \pm 0.001$ & $0.191 \pm 0.001$\\
       \textsc{Bert-large} & 336M & $55.2\% \pm 1.2\%$ & $0.116 \pm 0.002$ & $0.226 \pm 0.003$ & $38.9\% \pm 2.9\%$ & $0.1 \pm 0.001$ & $0.191 \pm 0.001$\\
    \midrule
       Null \textsc{Gpt-2} & 124M & $27.3\% \pm 1\%$ & $0.209 \pm 0.003$ & $0.322 \pm 0.005$ & $13.8\% \pm 1.7\%$ & $0.145 \pm 0.001$ & $0.248 \pm 0.003$\\
    \bottomrule
    \end{tabular}}
    \label{tab:20ng}
\end{table*}

\subsubsection{Topic distribution can be recovered when LDA is a sequence sub-component}

Most natural texts are not exchangeable. For example, consider this sentence: ``The \textbf{sediment} found in the \textbf{quartz} includes \textbf{silicon}.'' This sentence is neither fully exchangeable nor generated purely by a latent state model. However, if we only consider the words that contribute to the sentence’s topic (geology), we are left with \textbf{sediment}, \textbf{quartz}, \textbf{silicon}. These words can be plausibly generated by an exchangeable topic model as a sub-component: the sentence ``The \textbf{silicon} found in the \textbf{sediment} includes \textbf{quartz}'' also has the topic of geology even though word order changes. While we do not exhaustively list possible factors behind language that can be embedded by LLMs, we study to what extent can topics be encoded by model embeddings under settings where topical words occupy a varied proportion with respect to the whole sequence.

\paragraph{Model} Similar to the LDA experiments, we generate synthetic documents with an HMM-LDA model \citep{NIPS2004_ef0917ea} (Figure \ref{fig:hmm-lda}). HMM-LDA combines semantics and syntax, and posits that each word comes from a latent class and that class transitions are governed by an HMM. Using the HMM-LDA model allows one to manipulate the degree to which topic exists as a sub-component of the sequence. 

When $p$ is 1, all the words are generated from the LDA model. As $p$ decreases, more words are generated from the syntactic classes.  We can then examine how well topic distributions are recovered from the embeddings of models that are trained on data generated from HMM-LDA models that vary $p$.

\paragraph{Results} Results in Figure \ref{fig:hmm-lda} suggest that without exchangeability, the autoregressive transformer still learns the same semantic latent variables provided there is a large enough proportion of LDA-generated component (i.e.,~$p\geq0.4$). This suggests that the theory of finding explainable quantities in autoregressive models can potentially be extended to real texts where interpretable probabilistic models, such as the topic model, exist as a sub-component of language. 

% \begin{table}[h]
% \centering
% \caption{Topic prediction performance of the autoregressive transformer on HMM-LDA generated synthetic datasets. \textit{The language model successfully encodes semantics if the semantic part occupies a considerable enough proportion.}}
% \resizebox{0.5\columnwidth}{!}
% {\begin{tabular}{cccc} 
%  \toprule
%  & $\gamma=0.01$ & $\gamma=0.5$ & $\gamma=10$ \\
%  \midrule
%  $p = 1$ & $90.8\% \pm 0.4\%$ & $90.1\% \pm 1.2\%$ & $91.0\% \pm 0.8\%$ \\
%  $p = 0.9$ & $89.8\% \pm 0.7\%$ & $90.4\% \pm 0.5\%$ & $89.5\% \pm 0.6\%$ \\
%  $p = 0.8$ & $89.6\% \pm 1.5\%$ & $89.4\% \pm 1.0\%$ & $89.9\% \pm 1.8\%$ \\
%  $p = 0.5$ & $83.9\% \pm 1.9\%$ & $84.0\% \pm 0.9\%$ & $83.9\% \pm 1.7\%$ \\
%  $p = 0.4$ & $81.2\% \pm 0.8\%$ & $80.7\% \pm 2.7\%$ & $79.1\% \pm 1.0\%$ \\
%  $p = 0.25$ & $36.8\% \pm 21.6\%$ & $20.7\% \pm 0.3\%$ & $20.1\% \pm 0.9\%$ \\
%  \bottomrule
% \end{tabular}}
% \label{tab:hmm-lda}
% \end{table}

\subsubsection{Recovering latent topic distributions in natural corpora}

This section turns to analyzing LLMs pretrained on natural language. 
Although an exhaustive list of predictive sufficient statistic is difficult to identify, topical words do form a proportion of factors underlying the text. This allows us to hypothesize that topic mixtures can be decoded from the transformer residual streams.
%Recognizing this complexity, we explore the nuanced ways in which certain facets of language might be treated differently. It may be plausible to regard words as serving either a syntactic or a semantic function \citep{NIPS2004_ef0917ea}.
%specific aspects or levels of language data as partially exchangeable \citep{diaconis1980finetti}. In partial exchangeability, subsets of random variables are exchangeable within themselves, but not across subsets. 
%In localized contexts, word order may not matter. For example, the sequences ``strawberry, banana, cherry'' and ``banana, cherry, strawberry'' could be semantically equivalent when describing a smoothie. Within such contexts, the words are partially exchangeable. At a more abstract level, text segments that represent discrete semantic units (e.g., sentences within a paragraph) might also be considered partially exchangeable if they contribute independently to the overall meaning. 

\paragraph{Datasets} We use \textit{20Newsgroups} (20NG) and WikiText-103 \citep{merity2016pointer}. 20NG is a collection of 18,000 posts written in a style similar to informal emails, divided into 20 subjects. Contrasting with the informal language style in 20NG, WikiText-103 consists of over 100 million tokens sourced from the set of verified articles on Wikipedia that are classified as Good and Featured. 

\paragraph{Setup} We train LDA models across three random seeds on each dataset, and use pretrained large language models as our LLM. Specifically, the LLMs are \textsc{Gpt-2}, \textsc{Gpt-2-medium}, and \textsc{Gpt-2-large} \citep{radford2019language}; \textsc{Llama 2} and \textsc{Llama 2-chat} \citep{touvron2023llama}; and \textsc{Bert}, and \textsc{Bert-large} \citep{devlin-etal-2019-bert}. The probes target LDA-learned topic mixtures with $K \in \{20, 100\}$. For the models, we additionally include a randomly initialized \textsc{Gpt-2}, called Null \textsc{Gpt-2}, as a control that can differentiate the contribution of the LLM from the contribution of the probe. Considering that the first token may lead to a useful embedding that stands for the $\langle \text{CLS} \rangle$ token in \textsc{Bert} series, we searched across the first, last, and average embeddings. 
%The performance metrics are the same as those used in the synthetic datasets of Section \ref{sec:4.1}.

\paragraph{Topic prediction results} Performance is shown in Table \ref{tab:20ng} for 20NG and in Table \ref{tab:wiki} in the Appendix for Wikitext-103. To predict mixtures of twenty topics, random guessing would yield a $5\%$ accuracy for $K=20$ or $1\%$ accuracy for $K=100$. The best-performing LLM on each dataset demonstrates success at encoding topic distributions by achieving $88.5\% / 74.2\%$ accuracy on WikiText-103, and $62.9\% / 43.2\%$ on 20NG. The probes here take averaged token embeddings on the last layer in each LLM. While having direct information over whole sequences is advantageous compared to over an arbitrary token, we show that using only the last token preserves strong performance, with $73.7\% / 58.9\%$ accuracy on WikiText-103, and $52.2\% / 34.7\%$ on 20NG (Table \ref{tab:wiki-lasttoken} and \ref{tab:20ng-lasttoken} in Appendix). These results suggest that topic mixtures are directly encoded in embeddings, rather than resulting from high-quality word representations.

\paragraph{Topics encoded by inner layers} We also explored the possibility of decoding topic mixtures from other LLM layers, as opposed to only the last layer (Table \ref{tab:layers} in Appendix \ref{sec:appendix_results}). In the \textsc{Llama-2} series, the decodability of topic mixtures progressively increases as we move from the word embedder to the intermediate layers. Although a word embedder can potentially achieve high accuracy as demonstrated by the synthetic dataset, we observe that \textsc{Llama-2} relies on its inner layers to encode topic mixtures. This phenomenon can be caused by \textsc{Llama-2}'s stronger ability to model the next word, which is correlated with the ability to capture latent structures, which we discuss in more detail in the next paragraph.
% This phenomenon contrasts with \textsc{Bert}, where the performance of its later layers cannot be clearly distinguished from that of its word embedder.

\paragraph{Probe performance and LLM perplexity} To show that the the posterior predictive on topics corresponds to autoregressive prediction, we analyze 100 probes trained on 100 different token positions, along with their corresponding LLM perplexity (Figure \ref{fig:token-acc-perp}).  Each position is defined based on the corresponding percentile of the total document (each document has a different length). We expect that as perplexity on a token increases, probe performance based on the embedding taken from that token would decrease. This hypothesis is supported by the linear trend in Figure \ref{fig:token-acc-perp}. 
A possible confound in this result is that perplexity tends to decrease as the position in a document increases, so the results could be driven by position rather than perplexity; however, we ran a linear mixed-effects model and found that perplexity continued to have a statistically significant effect controlling for position (see Appendix \ref{sec:appendix_lmer}).

\section{Discussion}
\label{sec:disc}

In this paper, we have shown that autoregressive modeling is connected to Bayes optimal agents, where the sufficient statistic is encoded by the model in two major cases: 1) exchangeable distributions that emcompass a wide range of generative models including the topic model; and 2) latent state models that include the HMM. We have also seen that when these latent variables form a significant subcomponent of the data, they can still be encoded, leading to an explanatory approach on why topic mixtures are embedded in transformer residual streams.

There are several potential implications and future directions of our findings.

\paragraph{Interpretability.} Understanding the inner workings of LLMs is important for AI safety and trustworthiness. However, the sheer size of LLMs makes it challenging to analyze. Our results suggest that interpretable models of document structure - such as topic models - provide useful guidance about what mechanistic interpretability should look for. This in turn means that one goal for interpretability work should be enhancing models like LDA because such advances will in turn sharpen our ability to interpret LLM representations. 

\paragraph{Reverse direction - training LLMs to capture inductive biases.} We have shown how autoregressive models capture latent structures when predicting the next token. Another direction to explore is whether it is possible to explicitly incorporate inductive biases during training to more reliably and efficiently capture these latent structures, as have been done in modeling language \citep{mccoy-bayesian-prior2025} and vision \citep{carballo-concept-enhanced-2024}. 

\paragraph{Extensions to masked language models.} It has been shown that the autoregressive objective can be broken down into the Bayesian predictive distribution in several cases. It is possible to extend this breakdown to masked language models as well (Appendix \ref{sec:appendix_proof}). Whereas the autoregressive model learns a different posterior distribution $p(\theta | x_{1:n})$ for each $n$ as it progresses through tokens in a sequence, the masked model learns that same $p(\theta | x_{U})$, where $U$ denotes the set of unmasked indices, resulting in a less expressive objective. On synthetic topic model bags-of-words, the masked model underperforms, but it is worth exploring whether its performance is consistently worse than that of the autoregressive model in a wider range of cases.

\paragraph{Applications.} Our results help inform construction and use of embeddings in practice, and can be applied to generative models that satisfy the conditions discussed by the paper. The analyses can be naturally extended to other latent variables that do not depend heavily on word order, such as the author type of the document \citep{andreas-2022-language} or the author’s sentiment \citep{radford2017learning}. 

As another example, our results can also inform time series modelling. A practitioner might have a time series dataset with underlying factors that are informed by human experts. If the practitioner wants to use deep autoregressive models to construct an embedding that contains a certain kind of information from the input, they may need to ensure that this information is a predictive sufficient statistic for the task.

\section{Conclusion}
\label{sec:conc}

% \paragraph{Limitations and future work} Several limitations of our analysis motivate future work.

% \textit{Extending to more complex Bayesian probabilistic models.} We focused on three cases that consist of a one-step generation process and a simple likelihood $p(x|z)$ that can be approximated by a linear layer. Our analysis can be extended to hierarchical models or likelihoods parameterized by, for instance, neural networks, while simultaneously exploring more choices of transformer embeddings.

% \textit{Interpreting neural networks with Bayesian probabilistic models.} A similar Bayesian approach could be used to interpret the representations learned by other autoregressive neural networks. A direction for future work is extending our analyses to deep learning models for other modalities.

% \textit{Task difficulty.} As seen in the hypothesis space experiments, the ability to decode predictive sufficient statistics decreases as task difficulty increases. Developing a better understanding of the relationship between task difficulty and transformer embeddings is an important direction for future work.

We have developed a general framework for analyzing what the embeddings of an autoregressive language model should represent. Our analyses suggest that such embeddings should represent latent structures such that the next token $x_{n+1}$ is independent from previous tokens $x_{1:n}$ when conditioned on that structure, a property possessed by predictive sufficient statistics. We confirmed this hypothesis with probing experiments on cases where predictive sufficient statistics can be identified. We hope that our findings contribute to bridging the gap between Bayesian probabilistic models and deep neural networks.

% \subsubsection*{Broader Impact Statement}
% In this optional section, TMLR encourages authors to discuss possible repercussions of their work,
% notably any potential negative impact that a user of this research should be aware of. 
% Authors should consult the TMLR Ethics Guidelines available on the TMLR website
% for guidance on how to approach this subject.

% in many journals. This is optional and at the discretion of the authors. Only add
% this information once your submission is accepted and deanonymized. 

\subsubsection*{Acknowledgments}

This work was supported by ONR grant number N00014-23-1-2510. TRS is supported by an NDSEG Fellowship. RTM was supported by the NSF SPRF program under Grant No.\ 2204152.

\bibliography{main}
\bibliographystyle{tmlr}

\newpage
\appendix
\section{Appendix}

\subsection{Masked language models}
\label{sec:appendix_proof}

Masked language models (MLMs) are another class of LLMs that have been successful at modeling text, and we explore whether the objective of MLMs is also equivalent to implicit Bayesian inference under the exchangeability assumption. We find that it is, but it differs from autoregressive language models in a way that results in less expressivity.

MLMs, notably including the \textsc{Bert} series \citep{devlin-etal-2019-bert}, randomly mask certain tokens in the input sequence, and the model's goal is to predict the original words at the masked indices only by considering their surrounding context. Unlike autoregressive models, the MLM does not model a coherent joint distribution of the data \citep{yamakoshi-etal-2022-probing, young2022inconsistencies}. However, the log objective can be extended as follows,
\begin{align}
    \mathcal{L}_{MLM}(x_{1:N}) &= \sum_{n \in M} \log p(x_n | x_{i,i\in U}) \nonumber \\
    &= \sum_{n \in M} \log \int p(x_n | \theta)p(\theta|x_{i,i\in U})d\theta. \label{eq:mlm-bayes}
\end{align}
where $M$ denotes the set of masked indices, and $U$ denotes the set of unmasked indices. The proof is given below, and is a simple extension of the derivation for the autoregressive version. The difference between the MLM objective and the autoregressive objective is that in the summation, the prediction of each token $x_n$ uses the same posterior over the latent variable $p(\theta|x_{i,i\in U})$. In other words, each token $x_n$ is predicted independently from the latent variable $\theta$. As a result, MLM forms a less expressive Bayesian inference objective than autoregressive models. Therefore, we aim to empirically evaluate both the ability of MLM to recover latent variables, and whether its performance differs from that of autoregressive models.

%TLG: at the end here we need to set up why we are going to focus on LDA for our target latent distributions

We now prove the equivalence between the MLM objective and Bayesian inference in Equation \ref{eq:mlm-bayes}. The statement is that, given an exchangeable process $x_1, x_2, ..., x_n$,
\begin{align}
    \mathcal{L}_{MLM}(x_{1:N}) &:= \sum_{n \in M} \log p(x_n | x_{i,i\in U}) \\
    &= \sum_{n \in M} \log \int p(x_n | \theta)p(\theta|x_{i,i\in U})d\theta, \label{eq:mlm_bayes_appendix}
\end{align}
where $M$ is the set of masked indices and $U$ is the set of unmasked indices, and by construction $M \cap U = \emptyset$.

\paragraph{Proof.} We first prove the autoregressive version of the equivalence, which \citet{korshunova2018bruno} proposes for each individual term in the summation but briefly mentions why it is equivalent,
\begin{align}
    \log p(x_{1:N}) &= \log p(x_1) + \sum_{n=1}^{N-1} \log p(x_{n+1}|x_{1:n}) \nonumber \\
    &= \log p(x_1) + \sum_{n=1}^{N-1} \log \int_\theta p(x_{n+1}|\theta)p(\theta|x_{1:n})d\theta.
\end{align}
To do so, we shall prove the equivalence in each term in the summation, that is, the statement that
\begin{align}
    p(x_{n+1}|x_{1:n}) = \int_\theta p(x_{n+1}|\theta)p(\theta|x_{1:n})d\theta \label{eq:gpt_bayes_appendix}
\end{align}
for each $n>1$. First, de Finetti's theorem states that, under the same exchangeability condition,
\begin{align}
    p(x_{1:n+1}) = \int_{\theta} p(\theta)\prod_{i=1}^{n+1}p(x_i|\theta) d\theta,
\label{eq:definetti-appendix}
\end{align}
and also that each $x_i$ is conditionally independent given $\theta$ for all $i$. To show the equivalence, we first divide each side of Equation \ref{eq:definetti-appendix} by $p(x_{1:n})$, assuming that $p(x_{1:n}) > 0$. The left hand side becomes,
\begin{align}
    \frac{p(x_{1:n+1})}{p(x_{1:n})} = p(x_{n+1}|x_{1:n}).
\end{align}
The right hand side becomes,
\begin{align}
\int_{\theta} p(\theta)\frac{\prod_{i=1}^{n+1}p(x_i|\theta)}{p(x_{1:n})} d\theta &= \int_{\theta} p(\theta)\frac{p(x_{n+1}|\theta) p(x_{1:n}|\theta)}{p(x_{1:n})} d\theta \label{eq:001}\\
&= \int_{\theta} p(x_{n+1}|\theta) \frac{p(\theta)p(x_{1:n}|\theta)}{p(x_{1:n})} d\theta \\
&= \int_{\theta} p(x_{n+1}|\theta) p(\theta|x_{1:n} )d\theta. \label{eq:002}
\end{align}
Line \ref{eq:001} uses conditional independence to combine the product on $i=1$ through $n$, and line \ref{eq:002} uses Bayes rule. Therefore, because of Equation \ref{eq:definetti-appendix}, we prove the statement of Equation \ref{eq:gpt_bayes_appendix}.

The proof for MLM in Equation \ref{eq:mlm_bayes_appendix} can be shown by a simple extension. We shall use $x_U$ as short hand for $x_{i,i\in U}$. Using de Finetti's theorem, 
\begin{align}
    p(x_{\{n\} \cup U}) = \int_{\theta} p(\theta)\prod_{i \in \{n\} \cup U}p(x_i|\theta) d\theta,
\label{eq:definetti-appendix-mlm}
\end{align}
Dividing each side of the above equation by $p(x_U)$, the left hand side becomes,
\begin{align}
    \frac{p(x_{\{n\} \cup U})}{p(x_U)} = p(x_n|x_U).
\end{align}
The right hand side becomes,
\begin{align}
\int_{\theta} p(\theta)\frac{\prod_{i \in \{n\} \cup U}p(x_i|\theta)}{p(x_U)} d\theta &= \int_{\theta} p(\theta)\frac{p(x_{n}|\theta) p(x_U|\theta)}{p(x_U)} d\theta \label{eq:003}\\
&= \int_{\theta} p(x_{n}|\theta) \frac{p(\theta)p(x_{U}|\theta)}{p(x_{U})} d\theta \\
&= \int_{\theta} p(x_{n}|\theta) p(\theta|x_{U} )d\theta. \label{eq:004}
\end{align}
Same as in the proof for the autoregressive version, line \ref{eq:003} uses conditional independence to combine the product on $U$, and line \ref{eq:004} uses Bayes rule. Therefore, we have that $p(x_n | x_{i,i\in U}) = \int p(x_n | \theta)p(\theta|x_{i,i\in U})d\theta$.
Thus, we have shown the equivalence for each term in the summation of the original statement Equation \ref{eq:mlm_bayes_appendix}.

\subsection{Definition of additional exchangeable conjugate models}
\label{sec:appendix-models}
We consider the generative process for sequence $x_i$, where $x_{ij}$ are i.i.d. across $j$.

\subsubsection{Topic model}

The generative model for LDA is

\begin{tabularx}{\linewidth}{@{}p{.4\linewidth}p{.6\linewidth}@{}}
    For each topic $k$ in $(1,...,K),$
    \begin{enumerate}
        \item Draw topic $\beta_k \sim \text{Dirichlet}_V(\eta)$.
    \end{enumerate} 
    &
    For each document $i$,
        \begin{enumerate}
            \item Draw topic mixture $\theta_i \sim \text{Dirichlet}(\alpha).$
            \item For each word $j$ in document $i$,
            \begin{enumerate}
                \item Draw topic assignment $t_{ij} \sim \text{Categorical}(\theta_i)$,
                \item Draw word $x_{ij} \sim \text{Categorical}(\beta_{t_{ij}})$,
            \end{enumerate}
        \end{enumerate}
\end{tabularx}

where $V$ is the vocabulary size, and $\alpha$ and $\eta$ are pre-initialized hyperparameters. 

\subsubsection{Beta-Bernoulli model}
\begin{align*}
    \theta_i &\sim \textsc{Beta}(\alpha, \beta) \\
    x_{ij} &\sim \textsc{Bernoulli}(\theta_i),
\end{align*}
where $\alpha, \beta$ are fixed hyperparameters.

\subsubsection{Gamma-Exponential model}
\begin{align*}
    \theta_i &\sim \textsc{Gamma}(\alpha, \beta) \\
    x_{ij} &\sim \textsc{Exponential}(\theta_i),
\end{align*}
where $\alpha, \beta$ are fixed hyperparameters.

\subsubsection{HMM-LDA}

There are $C$ classes $\{c_1,...,c_C\}$ which follow an HMM transition matrix. One class is the semantic class $c_1$, and a word coming from $c_1$ is generated by an LDA model. Words of other classes are ‘syntactic’ and are directly generated by a distribution over words with a Dirichlet prior, one distribution for each $\{c_2,...,c_C\}$. The probability of generating from $c_1$, the LDA class, is determined by a parameter $p$ which specifies the transition probability into that class from all other classes. The remaining transition probabilities are generated from a Dirichlet($\gamma$) prior; lower $\gamma$ corresponds to more sparse transition matrices. 

\subsection{Implementational details}
\label{sec:appendix-implementation}

All computations for synthetic datasets are run on single Tesla T4 GPUs, and those for natural corpora are run on single A100 GPUs. 

\paragraph{Experimental process} Each dataset is split into three sets: set 1, set 2, and set 3. Set 1 is used for training the transformer. Set 2 is used for validating the transformer and getting embeddings from transformer that are used to train the probe. Set 3 is used for validating the probe.

Except discrete hypothesis space datasets and natural corpora, the sizes for the three sets are: 10000, 3000, 1000, and each sequence is 500-tokens long. In the discrete hypothesis space datasets, we experimented with different sequence lengths (detailed in our results), and the sizes for the three sets are: 20000, 19000, 1000. 

In HMM-LDA, sequence lengths are 400, and the sizes for the three sets are 10000, 1000, 1000.

On 20NG, probe training and validation are run on 11,314 and 7,532 documents, respectively. On WikiText-103, probe training and validation are run on 28,475 and 60 documents, respectively. Both splits are derived directly from train-validation split provided by the dataset sources. Note that set 1 is not used for natural corpora because we use pretrained LLMs.

\paragraph{Transformer} Except for topic models, we use a three-layer transformer decoder with hidden-size $=128$ and number of attention heads $=8$. If the input is categorical (similar to tokens in natural corpus), we employ the standard word embedder layer before the decoder layers. If the input is continuous, we use a Linear layer to map inputs to dimension 128 in place of the word embedder layer.

Dropout $=0.1$ is applied, and learning rate $=0.001$, batch-size $=64$.

\paragraph{Transformers on topic models}

Autoregressive transformer (AT) and \textsc{Bert} hyperparameters for training are given in Table \ref{tab:syn-appendix-1}. Configurations of \textsc{Bert} are identical to those of \textsc{Bert-tiny} from \citet{Turc2019WellReadSL}. AT hidden sizes and final layer embedding sizes are 128, same as \textsc{Bert-tiny}, and it uses four layers, resulting in 655,336 parameters. \text{Bert} has 608,747 parameters.

\paragraph{Probe} The probe is a linear layer with softmax activations. Learning rate is tuned in $[0.001, 0.01]$, and batch-size $=64$.

\paragraph{Hyperparameters for exchangeable conjugate models}

The Gaussian-Gamma hyperparameters $\alpha_0, \beta_0, \mu_0, \lambda_0$ are $\{5,1,1,1\}$. The OOD hyperparameters are $\{2,1,5,1\}$.

On the Beta-Bernoulli model, we use $\alpha = 2, \beta = 8$. In the OOD case, $\alpha = 8, \beta = 2$.

On the Gamma-Exponential model, we use $\alpha = 2, \beta = 4$. In the OOD case, $\alpha = 2, \beta = 1$.

\subsection{Linear mixed-effects model}
\label{sec:appendix_lmer}

We want to validate our hypothesis that the LLM's latent topic representation helps it predict individual tokens. While Fig.~\ref{fig:token-acc-perp} is suggestive of this relationship, here we use statistical testing to confirm it.

Concretely, we use a linear mixed-effects model to predict the per-token perplexity. We analyze 701,243 individual tokens from 20NG test corpus using \textsc{Gpt-2}. Perplexity naturally decreases as the LLM processes the document, so we include a fixed effect of token position and a random effect for the document itself; finally, we include the topic decoding accuracy (a binary 0 / 1 outcome based on the topic probe) as the variable of interest. We extract 100 tokens per document, stratified so they are evenly spaced, and represent the token position as the percent into the document,

\begin{equation}
    \text{perplexity} \sim \text{token\_position} + \text{topic\_accuracy} + (1 | \text{document\_id}).
\end{equation}

We find significant effects for both token position and topic accuracy,

\begin{table}[ht]
\centering
\resizebox{0.9\columnwidth}{!}{\begin{tabular}{lllrrrrr}
  \hline
Effect & Group & Term & Estimate & Std. Error & Statistic & DF & p-value \\ 
  \hline
fixed &  & (Intercept) & 4.65 & 0.01 & 413.28 & 21078.63 & $<$2e-16 \\ 
fixed &  & topic\_accuracy & -0.15 & 0.01 & -16.51 & 355354.64 & $<$2e-16 \\ 
fixed &  & token\_position & -0.78 & 0.01 & -58.47 & 696289.69 & $<$2e-16 \\ 
ran\_pars & document\_id & sd\_\_(Intercept) & 0.63 &  &  &  &  \\ 
ran\_pars & Residual & sd\_\_Observation & 3.22 &  &  &  &  \\ 
   \hline
\end{tabular}}
\end{table}

Finally, we obtain a Variance Inflation Factor of 1.014742 between accuracy and token position, suggesting an acceptable degree of colinearity between the two variables. 

\subsection{Additional results}
\label{sec:appendix_results}

\paragraph{Hyperparameter sweep}

Table \ref{tab:hp} shows hyperparameter sweep in the Bayesian conjugate models setting, across transformer embedding size in $\{8, 32, 128\}$, number of layers in $\{2, 3, 4\}$, and number of attention heads in $\{4, 8\}$. In general, we observe that embedding size affects performance most significantly. In the Gaussian-Gamma and Bernoulli datasets, performance improves with higher embedding size. In the Gamma-Exponential dataset, performance is best with embedding size $= 32$.

Hyperparameters for probes on AT and \textsc{Bert} on LDA are given in Table \ref{tab:syn-appendix-2}.

Hyperparameters for probes on the LLMs in natural corpora are given in Table \ref{tab:nat-appendix-1} and Table \ref{tab:nat-appendix-2}. 

\begin{table}
\centering
\caption{Scaled MSE across hyperparameter settings where the probe targets sufficient statistics, along with standard error across three random seeds. In the Gaussian-Gamma case, we report only on the second sufficient statistic, i.e., the standard deviation of the seen sequence (which is more challenging than the mean), to avoid clutter.}
\subcaption{Gaussian-Gamma.}
\resizebox{0.8\columnwidth}{!}
{\begin{tabular}{ccccccc} 
 \toprule
 &&num heads $=4$&&&num heads $=8$& \\
 \midrule
 & $d=8$ & $d=32$ & $d=128$ & $d=8$ & $d=32$ & $d=128$ \\
 \midrule
 2 layers & $0.168 \pm 0.022$ & $0.076 \pm 0.023$ & $0.069 \pm 0.013$ & $0.303 \pm 0.039$ & $0.045 \pm 0.010$ & $0.050 \pm 0.005$\\
 3 layers & $0.566 \pm 0.017$ & $0.066 \pm 0.018$ & $0.048 \pm 0.005$ & $0.321 \pm 0.014$ & $0.034 \pm 0.005$ & $0.046 \pm 0.010$\\
 4 layers & $0.451 \pm 0.059$ & $0.056 \pm 0.008$ & $0.047 \pm 0.009$ & $0.490 \pm 0.046$ & $0.081 \pm 0.006$ & $0.033 \pm 0.004$\\
 \bottomrule
\end{tabular}}
\vspace{7mm}
\subcaption{Beta-Bernoulli.}
\resizebox{0.8\columnwidth}{!}
{\begin{tabular}{ccccccc} 
 \toprule
 &&num heads $=4$&&&num heads $=8$& \\
 \midrule
 & $d=8$ & $d=32$ & $d=128$ & $d=8$ & $d=32$ & $d=128$ \\
 \midrule
 2 layers & $0.010 \pm 0.002$ & $0.002 \pm 0.000$ & $0.001 \pm 0.000$ & $0.007 \pm 0.002$ & $0.002 \pm 0.001$ & $0.001 \pm 0.000$\\
 3 layers & $0.008 \pm 0.002$ & $0.001 \pm 0.000$ & $0.000 \pm 0.000$ & $0.010 \pm 0.001$ & $0.001 \pm 0.000$ & $0.001 \pm 0.000$\\
 4 layers & $0.007 \pm 0.001$ & $0.001 \pm 0.000$ & $0.000 \pm 0.000$ & $0.010 \pm 0.002$ & $0.001 \pm 0.000$ & $0.000 \pm 0.000$\\
 \bottomrule
\end{tabular}}
\vspace{7mm}
\subcaption{Gamma-Exponential.}
\resizebox{0.8\columnwidth}{!}
{\begin{tabular}{ccccccc} 
 \toprule
 &&num heads $=4$&&&num heads $=8$& \\
 \midrule
 & $d=8$ & $d=32$ & $d=128$ & $d=8$ & $d=32$ & $d=128$ \\
 \midrule
 2 layers & $0.004 \pm 0.001$ & $0.001 \pm 0.000$ & $0.009 \pm 0.006$ & $0.002 \pm 0.001$ & $0.001 \pm 0.001$ & $0.067 \pm 0.006$\\
 3 layers & $0.003 \pm 0.001$ & $0.001 \pm 0.000$ & $0.006 \pm 0.004$ & $0.002 \pm 0.000$ & $0.001 \pm 0.000$ & $0.058 \pm 0.006$\\
 4 layers & $0.002 \pm 0.000$ & $0.000 \pm 0.000$ & $0.021 \pm 0.008$ & $0.002 \pm 0.000$ & $0.001 \pm 0.000$ & $0.033 \pm 0.005$\\
 \bottomrule
\end{tabular}}
\label{tab:hp}
\end{table}

\begin{figure}[t]
    \centering
    \begin{subfigure}{0.495\textwidth}
        \centering
        \includegraphics[width=\textwidth]{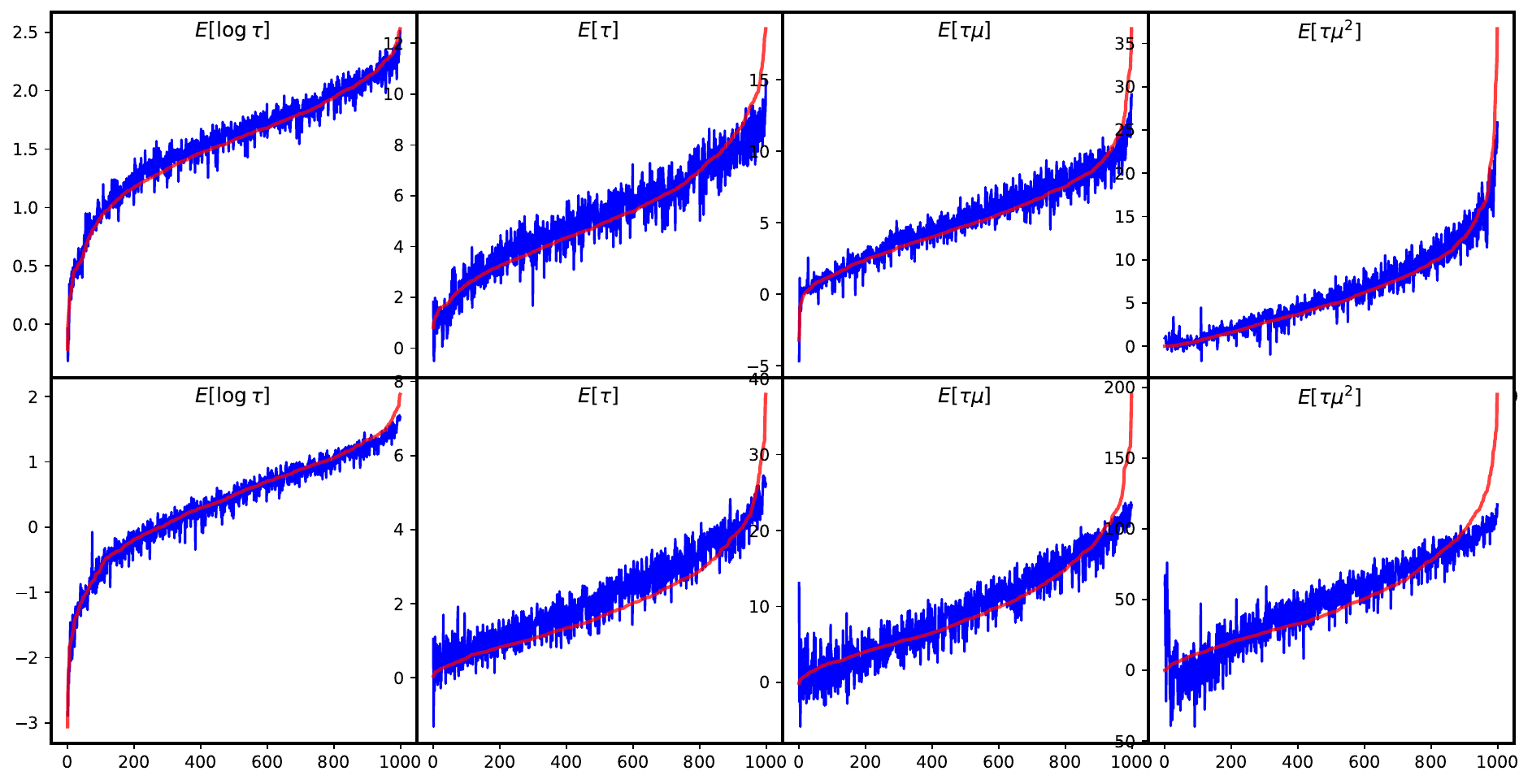}
        \caption{Gaussian-Gamma (linear probe).}
        \label{fig:gaussian-gamma-moment1}
    \end{subfigure}
    \begin{subfigure}{0.495\textwidth}
        \centering
        \includegraphics[width=\textwidth]{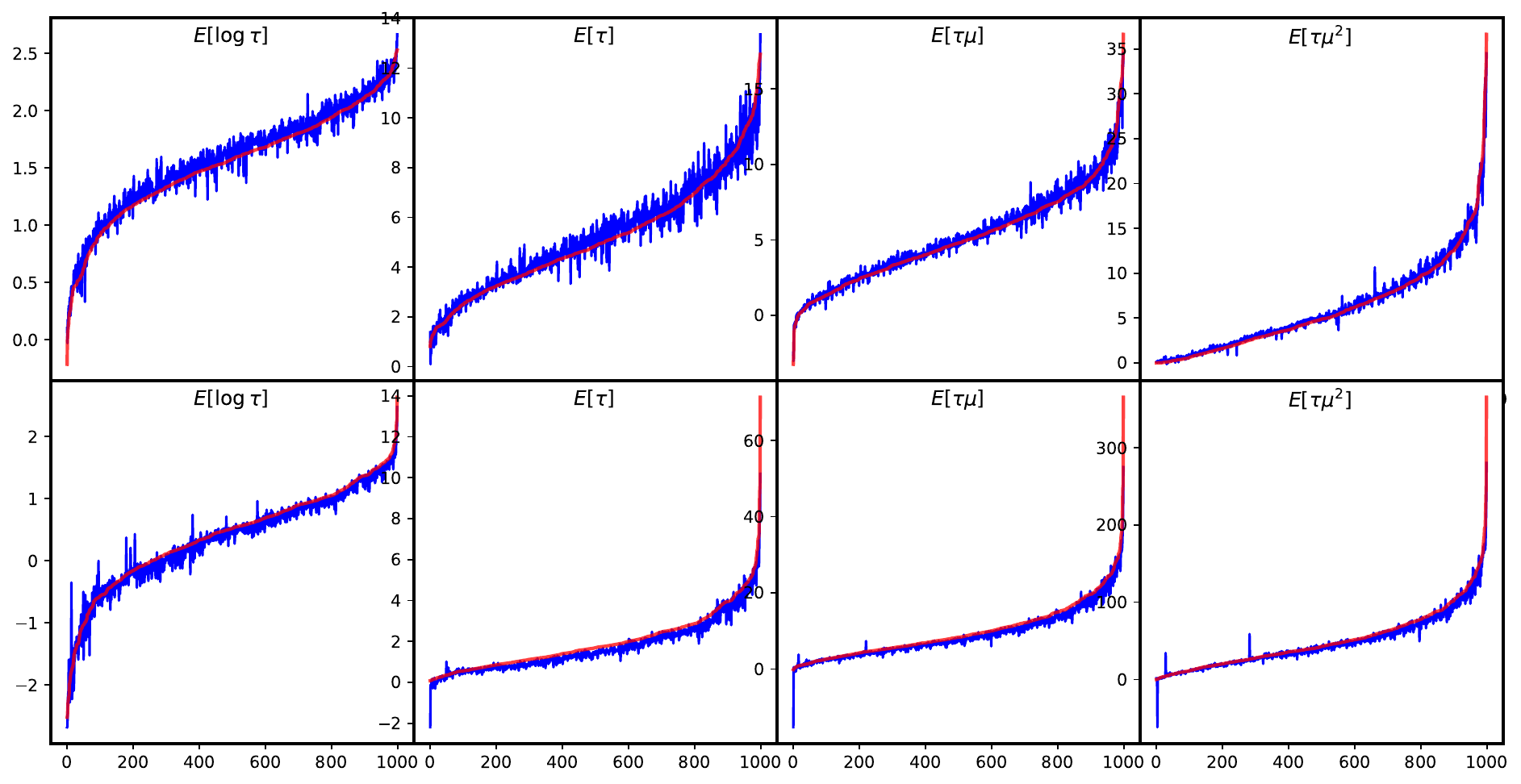}
        \caption{Gaussian-Gamma (two-layer probe).}
        \label{fig:gaussian-gamma-moment2}
    \end{subfigure}
    \vspace{-5mm}
    \caption{Probe recovery of transformer-learned posterior distribution moments (blue) and ground truth moments (red) across 1000 test datapoints. The first row shows parameters probed in the general non-OOD case. The second row shows corresponding information in the OOD case.}
    \label{fig:gaussian-gamma-moment}
\end{figure}

\paragraph{Moments} Figure \ref{fig:gaussian-gamma-moment1} shows that the probe decodes the moments of the posterior distribution of the Gaussian-Gamma model. Because the higher distribution moments are more volatile in value and less directly related to estimating the parameters $\theta$, we examine whether existing discrepancies are caused by an overly simple probe. We perform a second set of experiments where the probe has a hidden layer with ReLU activations (Figure \ref{fig:gaussian-gamma-moment2}), showing stronger alignment.

\begin{figure*}
    \centering
    \begin{subfigure}{0.65\textwidth}
        \centering
        \includegraphics[width=\textwidth]{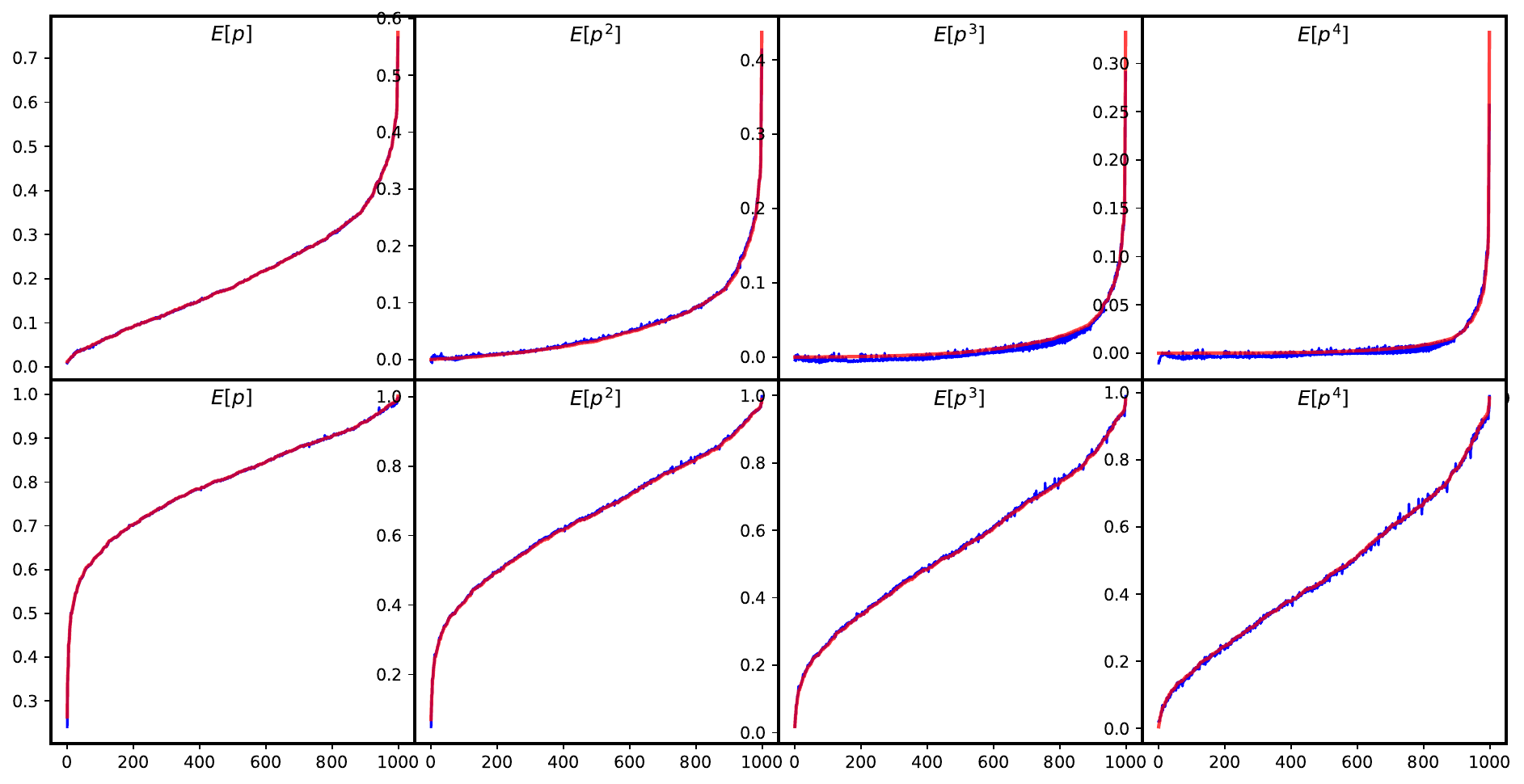}
        \caption{Beta-Bernoulli.}
    \end{subfigure}
    \begin{subfigure}{0.65\textwidth}
        \centering
        \includegraphics[width=\textwidth]{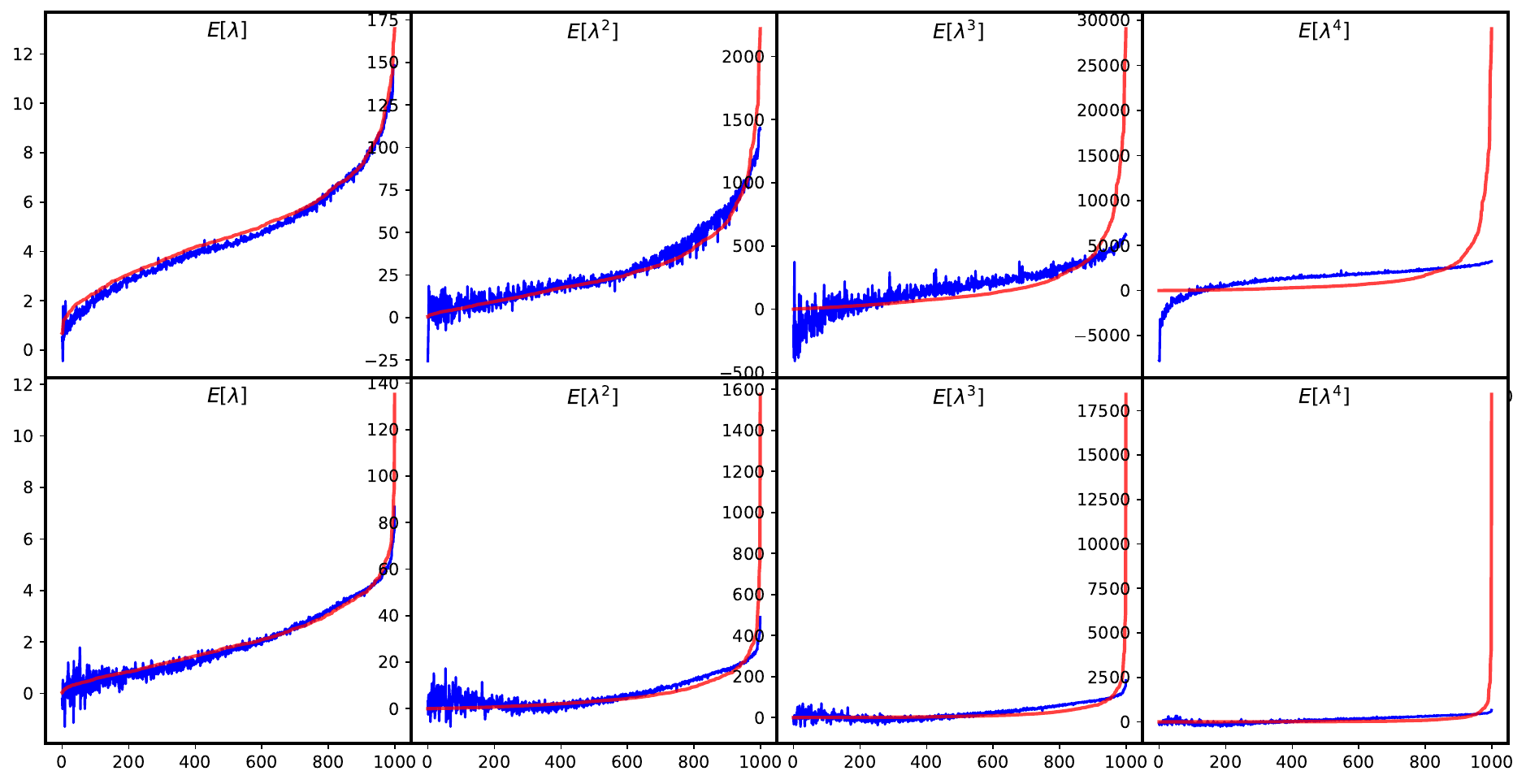}
        \caption{Gamma-Exponential.}
    \end{subfigure}
    \begin{subfigure}{0.65\textwidth}
        \centering
        \includegraphics[width=\textwidth]{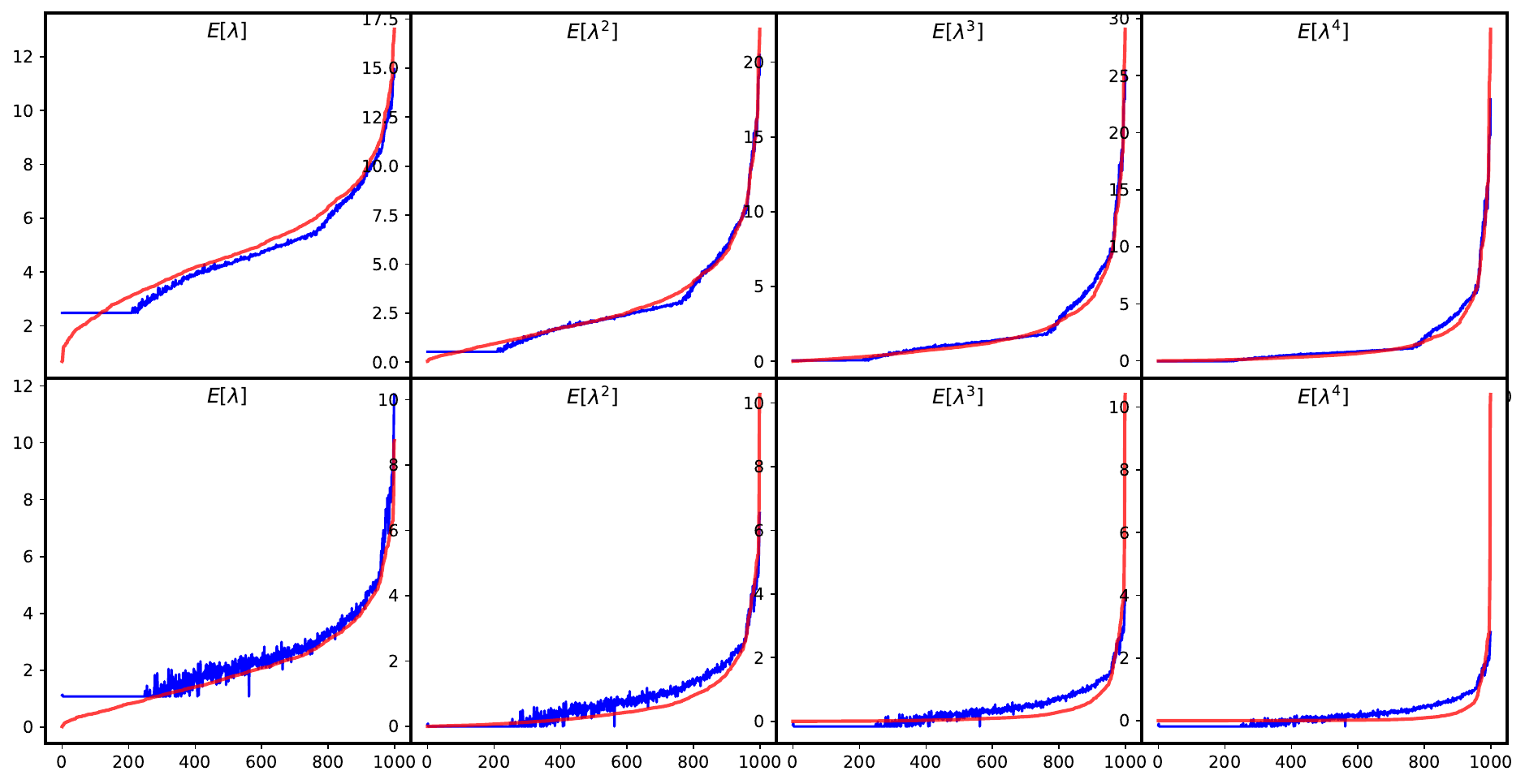}
        \caption{Gamma-Exponential (non-linear probe with factors).}
    \end{subfigure}
    \caption{Probe recovery of transformer-learned posterior distribution moments (blue) plotted with ground truth moments (red). The first row shows parameters probed on 1000 test datapoints in the general, i.e., non-OOD, case. The second row shows corresponding information in the OOD case.}
    \label{fig:conjugate-moment2}
\end{figure*}

Figure \ref{fig:conjugate-moment2} shows results on posterior distribution moments on Beta-Bernoulli and Gamma-Exponential models. On the Gamma-Exponential model, we divide the target second, third, and fourth moments by factors of 10, 100, 1000, respectively so that each moment is given roughly equal importance. 

\begin{figure}
    \centering
    \includegraphics[width=1\linewidth]{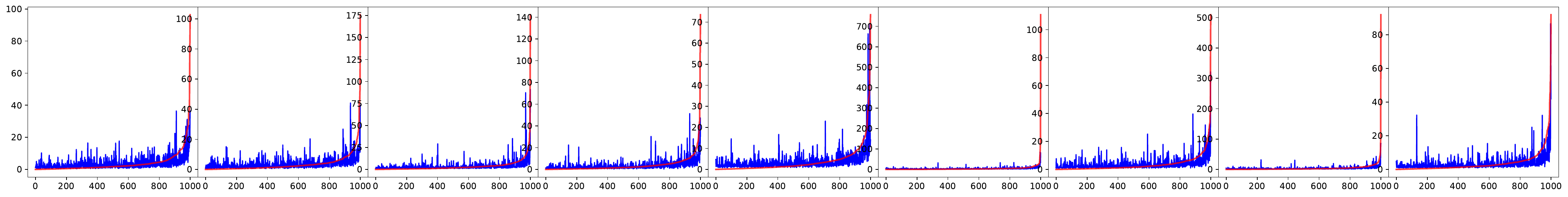}
    \caption{Probing over the first 10 tokens themselves using the 10th token embedding of the transformer.
Aside from perfectly encoding the 10th token, this embedding does not show memorization over the other 9
tokens as suggested by the noise in probe recovery. Visually, the noise is less than in the Gaussian-Gamma case (Figure \ref{fig:token-memorization}). This is partly due to spuriously large magnitude datapoints in this distribution, and the noise here is still significantly higher than in capturing sufficient statistics (Figure \ref{fig:conjugate-ss}).}
    \label{fig:memorization-exp}
\end{figure}

\paragraph{Memorization} Figure \ref{fig:memorization-exp} conducts the same memorization experiments for the Gamma-Exponential case as in the Gaussian-Gamma case in the main text (Figure \ref{fig:token-memorization}). 

\paragraph{Parsimonious embeddings} Table \ref{tab:parsimonious} shows results from training an MLP on sufficient statistics to target transformer embeddings. To compute these results, we get mean-squared error, which is then divided by the variance of the transformer embeddings along each dimension separately. The error is averaged across the embedding dimension and subtracted from 1 to yield the mean proportion of variance accounted for in the embedding. These results establish that the relationship runs in both directions: we can decode sufficient statistics from embeddings, and we can predict embeddings from sufficient statistics.

\begin{table*}[h!]
    \centering
    \caption{Validation variance of transformer embeddings accounted for by training an MLP only on sufficient statistics of training sequences. }
    \resizebox{0.5\columnwidth}{!}{
    \begin{tabular}{cccc}
    \toprule 
    Transformer size & Gaussian-Gamma	& Beta-Bernoulli	& Gamma-Exponential \\
    \midrule
    128	&71.1\%	&28.4\%	&72.2\%\\
    8	&84.6\%	&48.6\%	&79.6\%\\
    \bottomrule
\end{tabular}}
    \label{tab:parsimonious}
    \vspace{1mm}
\end{table*}

\begin{figure}
    \centering
    \includegraphics[width=0.4\textwidth]{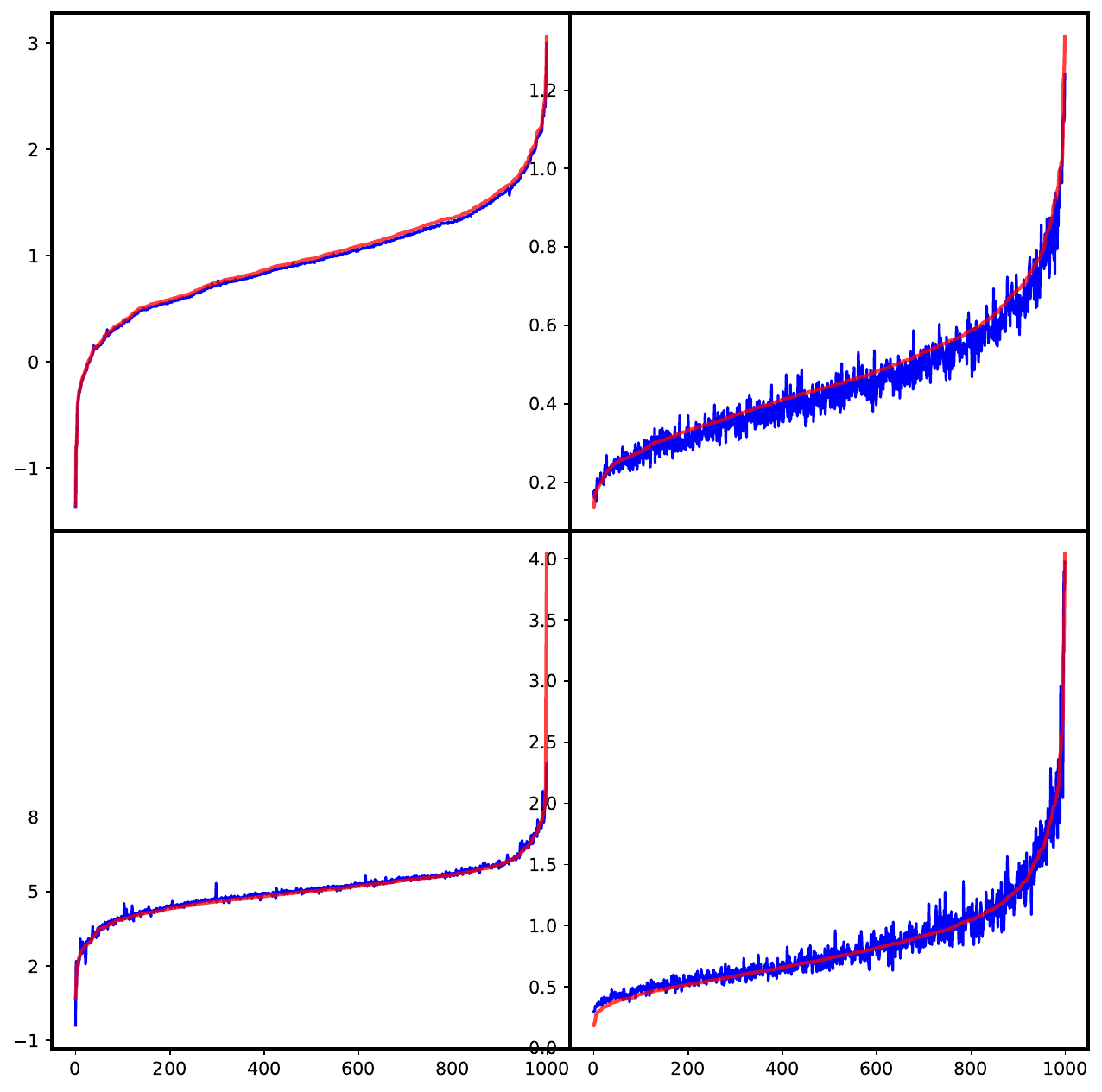}
    \caption{Probing over mean (left) and standard deviation (right) of the first 10 tokens using the 10th token embedding of the transformer in the Gaussian-Gamma dataset. The first row corresponds to the same generation process, and the second row corresponds to the OOD case.}
    \label{fig:token10}
\end{figure}

Figure \ref{fig:token10} shows Gaussian-Gamma probing results on sufficient statistics on the 10th token, complementing our exploration of token memorization in transformer (Figure \ref{fig:token-memorization}).

\paragraph{Probing results on Wikitext-103}

Table \ref{tab:wiki} shows probing results on Wikitext-103.

\begin{table*}
    \centering
    \caption{Wikitext-103 topic prediction performance based on different LLMs. \textit{On accuracy, the \textsc{Llama2-chat} performance is not statistically significantly different from that of the MLMs; otherwise, the autoregressive models statistically significantly outperform the non-autoregressive models.}}
    \resizebox{1\columnwidth}{!}{
    \begin{tabular}{cccccccc}
    \toprule 
    & & & $K=20$ & & & $K=100$ & \\ \cmidrule(lr){3-5} \cmidrule(lr){6-8} 
    Model & Parameters & Accuracy $\uparrow$ & L2 loss $\downarrow$ & Tot. var. loss $\downarrow$ & Accuracy $\uparrow$ & L2 loss $\downarrow$ & Tot. var. loss $\downarrow$ \\
    \midrule
       \textsc{Gpt-2}  & 124M & $86.7\% \pm 0.5\%$ & $0.025 \pm 0$ & $0.098 \pm 0$ & $73.9\% \pm 2.3\%$ & $0.026 \pm 0.001$ & $0.089 \pm 0.002$\\
       \textsc{Gpt-2-medium} & 355M & $88.2\% \pm 0.6\%$ & $0.024 \pm 0$ & $0.097 \pm 0.001$ & $74.2\% \pm 1.3\%$ & $0.025 \pm 0$ & $0.097 \pm 0.002$\\
       \textsc{Gpt-2-large} & 774M & $88.5\% \pm 0.8\%$ & $0.023 \pm 0$ & $0.094 \pm 0.001$ & $74.2\% \pm 1.4\%$ & $0.025 \pm 0$ & $0.088 \pm 0.001$\\
       \textsc{Llama 2} & 7B & $87.3\% \pm 1.7\%$ & $0.023 \pm 0$ & $0.091 \pm 0.001$ & $70.4\% \pm 1.1\%$ & $0.026 \pm 0$ & $0.09 \pm 0.001$\\
       \textsc{Llama 2-chat} & 7B & $85.3\% \pm 0.7\%$ & $0.024 \pm 0$ & $0.094 \pm 0$ & $69.9\% \pm 1\%$ & $0.026 \pm 0$ & $0.09 \pm 0$\\
    \midrule
       \textsc{Bert} & 110M & $84.9\% \pm 1.1\%$ & $0.027 \pm 0$ & $0.103 \pm 0.001$ & $72.4\% \pm 1.4\%$ & $0.029 \pm 0$ & $0.097 \pm 0.002$\\
       \textsc{Bert-large} & 336M & $85.4\% \pm 1.7\%$ & $0.03 \pm 0$ & $0.111 \pm 0$ & $72.1\% \pm 0.9\%$ & $0.031 \pm 0$ & $0.104 \pm 0$\\
    \midrule
       Null \textsc{Gpt-2} & 124M & $58.1\% \pm 1.8\%$ & $0.121 \pm 0.003$ & $0.247 \pm 0.006$ & $32.9\% \pm 3.2\%$ & $0.099 \pm 0.003$ & $0.195 \pm 0.007$\\
    \bottomrule
    \end{tabular}}
    \label{tab:wiki}
\end{table*}

\paragraph{Inner layer performance on natural corpora}
Table \ref{tab:layers} shows topic distribution recovery from LLM inner layers.

\begin{table*}[h!]
    \centering
    \caption{Across-layer performance in the format of (accuracy, L2-loss), where we move from the word embedder to the final layer (prior to word emission) from left to right.}
    \resizebox{1\columnwidth}{!}{
    \begin{tabular}{ccccccccc}
    \toprule 
    Model & Parameters & Layer 0 & Layer 1 & Layer 50\% & Layer 75\% & Layer 90\% & Layer -2 & Layer -1 \\
    \midrule
       \textsc{Gpt-2}  & 124M & $87.4\%, 0.027$ & $87.1\%, 0.025$ & $86.1\%, 0.025$ & $86.7\%, 0.026$ & $86.3\%, 0.026$ & $85.9\%, 0.026$ & 86.7\%, 0.025\\
       \textsc{Gpt-2-medium} & 355M & $88.2\%, 0.026$ & $86.9\%, 0.024$ & $85.5\%, 0.027$ & $85.1\%, 0.028$ & $83.9\%, 0.03$ & $81.8\%, 0.04$ & $88.2\%, 0.024$\\
       \textsc{Gpt-2-large} & 774M & $88.1\%, 0.026$ & $87.6\%, 0.024$ & $86.5\%, 0.025$ & $85.4\%, 0.026$ & $84.5\%, 0.028$ & $82.9\%, 0.03$ & $88.5\%, 0.023$\\
       \textsc{Llama 2} & 7B & $59.2\%, 0.192$ & $74.4\%, 0.104$ & $87.3\%, 0.025$ & $86.6\%, 0.023$ & $86.9\%, 0.024$ & $86.7\%, 0.024$ & $87.3\%, 0.023$\\
       \textsc{Llama 2-chat} & 7B & $58.8\%, 0.193$ & $74.9\%, 0.098$ & $87.2\%, 0.025$ & $87.5\%, 0.023$ & $86.5\%, 0.024$ & $85.8\%, 0.025$ & $85.3\%, 0.024$\\
    \midrule
       \textsc{Bert} & 110M & $88.4\%, 0.028$ & $88.6\%, 0.024$ & $86.2\%,0.028$ & $85.9\%, 0.03$ & $86.4\%, 0.028$ & $86.4\%, 0.028$ & $84.9\%, 0.027$\\
       \textsc{Bert-large} & 336M & $88.2\%, 0.029$ & $87.8\%, 0.026$ & $86.4\%, 0.027$ & $85\%, 0.03$ & $85.9\%, 0.029$ & $85.3\%, 0.03$ & $85.4\%, 0.03$\\
    \bottomrule
    \end{tabular}}
    \label{tab:layers}
\end{table*}

\begin{table}[]
    \centering
    \caption{Autoregressive transformer (AT) and \textsc{Bert} hyperparameters for training on the synthetic datasets.}
    \begin{tabular}{c c c}
    \toprule
        Parameter & Tuning range & Chosen value\\
    \midrule
        Batch-size & $[8,128]$ & $16$\\
        Learning rate & $[3\cdot 10^{-5}, 10^{-3}]$ & $10^{-4}$\\
    \bottomrule
    \end{tabular}
    \label{tab:syn-appendix-1}
\end{table}

\begin{table}[]
    \centering
    \caption{Probe hyperparameters for training on top of synthetic dataset language models.}
    \begin{tabular}{c c c}
    \toprule
        Parameter & Tuning range & Chosen value\\
    \midrule
        Batch-size & $[8,64]$ & $16$\\
        Learning rate & $[10^{-4}, 0.03]$ & $10^{-3}$\\
        Weight-decay & $[0, 3.4 \cdot 10^{-4}]$ & 0\\
        Embedding choice & $\{\text{First, Last, Average}\}$ & Last for AT / Average for \textsc{Bert}\\
    \bottomrule
    \end{tabular}
    \label{tab:syn-appendix-2}
\end{table}

\begin{table}[h!]
    \centering
    \caption{Probe hyperparameters for training on top of \textsc{Gpt-2}, \textsc{Gpt-2-medium}, \textsc{Gpt-2-large}, \textsc{Bert}, and \textsc{Bert-large}.}
    \begin{tabular}{c c c}
    \toprule
        Parameter & Tuning range & Chosen value\\
    \midrule
        Batch-size & $\{128\}$ & $128$\\
        Learning rate & $[10^{-5}, 10^{-3}]$ & $3 \cdot 10^{-4}$\\
        Weight-decay & $[0, 3.4]$ & $3.4 \cdot 10^{-3}$\\
        Embedding choice & $\{\text{First, Last, Average}\}$ & Average\\
    \bottomrule
    \end{tabular}
    \label{tab:nat-appendix-1}
\end{table}

\begin{table}[h!]
    \centering
    \caption{Probe hyperparameters for training on top of \textsc{Llama 2} and \textsc{Llama 2-chat}.}
    \begin{tabular}{c c c}
    \toprule
        Parameter & Tuning range & Chosen value\\
    \midrule
        Batch-size & $\{128\}$ & $128$\\
        Learning rate & $[10^{-5}, 10^{-3}]$ & $10^{-4}$\\
        Weight-decay & $[0, 3.4]$ & 0.34\\
        Embedding choice & $\{\text{First, Last, Average}\}$ & Average\\
    \bottomrule
    \end{tabular}
    \label{tab:nat-appendix-2}
\end{table}

\paragraph{Using only last token embedding on natural corpora}
Using averaged tokens as document embeddings performs better across models in both natural corpora. Here, we also report on results from using only last tokens as document embeddings (Table \ref{tab:20ng-lasttoken} and Table \ref{tab:wiki-lasttoken}).

\begin{table*}[h!]
    \centering
    \caption{20NG topic prediction performance based on different LLMs using the last token as document embedding.}
    \resizebox{0.95\columnwidth}{!}{
    \begin{tabular}{cccccccc}
    \toprule 
    & & & $K=20$ & & & $K=100$ & \\ \cmidrule(lr){3-5} \cmidrule(lr){6-8} 
    Model & Parameters & Accuracy $\uparrow$ & L2 loss $\downarrow$ & Tot. var. loss $\downarrow$ & Accuracy $\uparrow$ & L2 loss $\downarrow$ & Tot. var. loss $\downarrow$ \\
    \midrule
       \textsc{Gpt-2}  & 124M & $47.8\% \pm 1.8\%$ & $0.143 \pm 0.001$ & $0.26 \pm 0.002$ & $31\% \pm 3\%$ & $0.115 \pm 00.001$ & $0.217 \pm 0.002$ \\
       \textsc{Gpt-2-medium} & 355M & $47.6\% \pm 1.2\%$ & $0.144 \pm 0.001$ & $0.26 \pm 0.001$ & $30.8\% \pm 3\%$ & $0.116 \pm 0.001$ & $0.217 \pm 0.002$ \\
       \textsc{Gpt-2-large} & 774M & $47.8\% \pm 1.8\%$ & $0.144 \pm 0$ & $0.261 \pm 0.002$ & $31.1\% \pm 2.9\%$ & $0.116 \pm 0$ & $0.217 \pm 0.002$ \\
       \textsc{Llama 2} & 7B & $45.4\% \pm 2\%$ & $0.152 \pm 0$ & $0.267 \pm 0.001$ & $27.9\% \pm 2.9\%$ & $0.122 \pm 0.001$ & $0.223 \pm 0.003$\\
       \textsc{Llama 2-chat} & 7B & $46.1\% \pm 1.8\%$ & $0.15 \pm 0.001$ & $0.265 \pm 0.001$ & $28.5\% \pm 2.5\%$ & $0.122 \pm 0.001$ & $0.222 \pm 0.003$\\
    \midrule
       \textsc{Bert} & 110M & $51.7\% \pm 1.4\%$ & $0.126 \pm 0.002$ & $0.236 \pm 0.002$ & $34.9\% \pm 2.8\%$ & $0.108 \pm 0$ & $0.204 \pm 0.001$ \\
       \textsc{Bert-large} & 336M & $52.2\% \pm 2.1\%$ & $0.125 \pm 0.001$ & $0.235 \pm 0.002$ & $34.7\% \pm 2.7\%$ & $0.107 \pm 0.001$ & $0.201 \pm 0.001$ \\
    \bottomrule
    \end{tabular}}
    \label{tab:20ng-lasttoken}
\end{table*}

\begin{table*}[h!]
    \centering
    \caption{WikiText-103 topic prediction performance based on different LLMs using the last token as document embedding.}
    \resizebox{0.95\columnwidth}{!}{
    \begin{tabular}{cccccccc}
    \toprule 
    & & & $K=20$ & & & $K=100$ & \\ \cmidrule(lr){3-5} \cmidrule(lr){6-8} 
    Model & Parameters & Accuracy $\uparrow$ & L2 loss $\downarrow$ & Tot. var. loss $\downarrow$ & Accuracy $\uparrow$ & L2 loss $\downarrow$ & Tot. var. loss $\downarrow$ \\
    \midrule
       \textsc{Gpt-2}  & 124M & $69.2\% \pm 1.4\%$ & $0.079 \pm 0.003$ & $0.191 \pm 0.005$ & $50.9\% \pm 1.3\%$ & $0.055 \pm 0.001$ & $0.146 \pm 0.003$\\
       \textsc{Gpt-2-medium} & 355M & $70.7\% \pm 0.6\%$ & $0.077 \pm 0.002$ & $0.188 \pm 0.004$ & $51.6\% \pm 1.9\%$ & $0.056 \pm 0.001$ & $0.146 \pm 0.003$\\
       \textsc{Gpt-2-large} & 774M & $72\% \pm 0.6\%$ & $0.074 \pm 0.001$ & $0.184 \pm 0.003$ & $52.8\% \pm 2\%$ & $0.053 \pm 0.001$ & $0.143 \pm 0.002$\\
       \textsc{Llama 2} & 7B & $69.7\% \pm 2.3\%$ & $0.075 \pm 0.002$ & $0.187 \pm 0.003$ & $51.9\% \pm 1.4\%$ & $0.054 \pm 0.001$ & $0.144 \pm 0.003$\\
       \textsc{Llama 2-chat} & 7B & $71\% \pm 1.8\%$ & $0.073 \pm 0.001$ & $0.181 \pm 0.002$ & $54.2\% \pm 0.8\%$ & $0.052 \pm 0.002$ & $0.141 \pm 0.005$\\
    \midrule
       \textsc{Bert} & 110M & $73.7\% \pm 1.8\%$ & $0.067 \pm 0$ & $0.177 \pm 0.002$ & $58.9\% \pm 0.9\%$ & $0.05 \pm 0.002$ & $0.14 \pm 0.001$\\
       \textsc{Bert-large} & 336M & $63.9\% \pm 3\%$ & $0.1 \pm 0.001$ & $0.219 \pm 0.001$ & $50\% \pm 1\%$ & $0.065 \pm 0.002$ & $0.164 \pm 0.003$\\
    \bottomrule
    \end{tabular}}
    \label{tab:wiki-lasttoken}
\end{table*}

\end{document}